%% file: main.tex
\documentclass[runningheads]{llncs}
\input{preamble}

\begin{document}

\title{Infinite-Homography as Robust Conditioning\\for Camera-Controlled Video Generation} 

\titlerunning{InfCam: Infinite-Homography for Camera-Controlled Video Generation}

\author{Min-Jung Kim\thanks{Authors contributed equally to this work.}\orcidlink{0000-0003-3799-8225} \and
Jeongho Kim\samethanks\orcidlink{0000-0003-4058-8163} \and
Hoiyeong Jin\orcidlink{0009-0002-4653-8888} \\
\and
Junha Hyung\orcidlink{0009-0002-8244-3659} \and
Jaegul Choo$^{\dagger}$\orcidlink{0000-0003-1071-4835}\\
}

\authorrunning{M.~Kim and J.~Kim et al.}

\institute{
    KAIST \\
    \email{\{emjay73, rlawjdghek,
    hy.jin, sharpeeee, jchoo\}@kaist.ac.kr}\\
    $^{\dagger}$~Corresponding author.
}

\maketitle

\begin{figure*}[h]
    \centering
    \vspace{-15pt}
    \includegraphics[width=\textwidth]{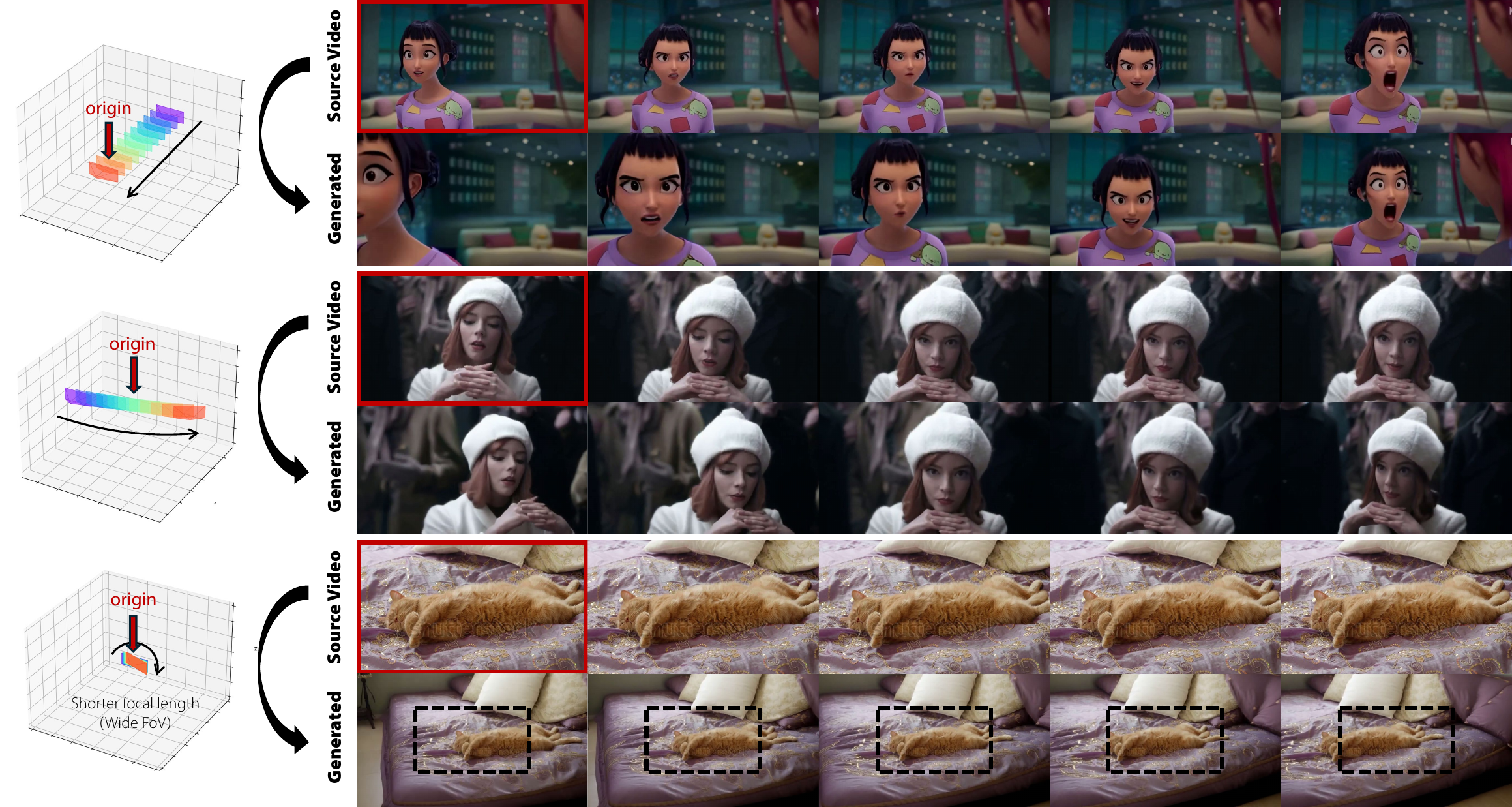}
    \captionof{figure}
    {\textbf{\modelname{} Results.} Given a video and a target camera trajectory, \modelname{} generates a video that faithfully follows the specified camera path.     
    The leftmost column visualizes the backward, arc, and rotational camera trajectories, 
    and the right side shows input–generated video pairs corresponding to each trajectory.
    The world coordinate origin is defined by the first frame's camera pose (highlighted in red).
    The rotational trajectory is generated with a shorter focal length to illustrate wide field-of-view generation.
    The black dashed box in the last row indicates the original field-of-view of the input video.
    }
    \vspace{-15pt}
    \label{fig:teaser}
\end{figure*}

\begin{abstract}
  Recent progress in video diffusion models has spurred growing interest in camera-controlled novel-view video generation for dynamic scenes, aiming to provide creators with 
cinematic camera control capabilities in post-production.
A key challenge in camera-controlled video generation is ensuring fidelity to the specified camera pose, while maintaining view consistency and reasoning about occluded geometry from limited observations.
To address this, existing methods either train trajectory-conditioned video generation model on trajectory–video pair dataset, 
or estimate depth from the input video to reproject it along a target trajectory and generate the unprojected regions.
Nevertheless, existing methods struggle to generate camera-pose–faithful, high-quality videos for two main reasons:
(1) reprojection-based approaches are highly susceptible to errors caused by inaccurate depth estimation; and 
(2) the limited diversity of camera trajectories in existing datasets restricts learned models.
To address these limitations, we present \modelname{}, a depth-free, camera-controlled video-to-video generation framework 
with high pose fidelity.
The framework integrates two key components:
(1) infinite homography warping, which encodes 3D camera rotations directly within the 2D latent space of a video diffusion model. Conditioning on this noise-free rotational information, the residual parallax term is predicted through end-to-end training to achieve high camera-pose fidelity; and
(2) a data augmentation pipeline that transforms existing synthetic multiview datasets into sequences with diverse trajectories and focal lengths.
Experimental results demonstrate that \modelname{} outperforms baseline methods in camera-pose accuracy and visual fidelity, generalizing well from synthetic to real-world data.
  \keywords{Camera Control \and Video Generation \and Infinite Homography}
\end{abstract}

\section{Introduction}
\label{sec:intro}
Changing the camera viewpoint in post-production is a highly sought-after video editing technique, as it eliminates the need for costly reshoots and enables creative visual effects. 
Recent advancements in large-scale video diffusion models have spurred research on controlling camera trajectories for novel-view video generation~\cite{bai2024syncammaster, van2024generative,ren2025gen3c, mark2025trajectorycrafter, bai2025recammaster}. 

However, existing methods still suffer from several limitations.
First, approaches that explicitly incorporate depth projection~\cite{mark2025trajectorycrafter, ren2025gen3c, zhang2025recapture} suffer from inherent performance degradation due to reprojection errors arising from inaccurately estimated depths.
While reprojection serves as an effective mechanism for aligning generated views with the target trajectory when accurate depth maps are available, it can also become a major source of error when the predicted depths are inaccurate.
Consequently, the overall performance of such systems remains fundamentally constrained by the accuracy and reliability of the underlying depth predictor.
Second, methods that learn the relationship between camera trajectory variations and novel-view video generation from trajectory–video pair datasets ~\cite{bai2025recammaster,bai2024syncammaster,van2024generative} tend to inherit dataset-specific biases. 
When the training data contain biased 
input–output video pairs, the network is likely to internalize these biases, leading to degraded generalization performance.
To address these limitations, we propose a robust infinite homography-based conditioning method.
Unlike existing depth reprojection-based methods that suffer from cascading errors due to their reliance on noisy depth estimates, 
our infinite homography warping module decomposes the reprojection process into known rotation and unknown translation components. 
We condition the model on the noise-free rotation component, while 
the residual translation component is learned end-to-end by leveraging strong prior knowledge from large video generation models.
Additionally, we introduce a data augmentation strategy that transforms datasets with constrained viewpoints into flexible trajectory formats 
while simultaneously augmenting focal lengths. 
By synthesizing diverse camera movements and field-of-view variations, our model learns robust view synthesis.
Extensive experiments demonstrate that the proposed infinite homography warping module and data augmentation enable 
superior trajectory fidelity compared to state-of-the-art methods. 
\cref{fig:teaser} demonstrates the generalizability of our approach on challenging real-world videos.
Our contributions are summarized as follows:
\begin{itemize}
\item We propose a camera-controlled novel-view video generation framework that achieves high pose fidelity without a depth prior.
\item We introduce an infinite homography warping module 
that conditions on a noise-free rotation term, 
predicting the residual parallax term through end-to-end training.
\item We present a data augmentation strategy that transforms existing datasets with limited viewpoints into diverse training pairs with flexible camera trajectories and varying focal lengths, enabling robust generalization
\item Extensive experiments demonstrate that our infinite homography warping and data augmentation strategy significantly improve camera-controlled video generation, achieving state-of-the-art performance on both synthetic and real-world videos.
\end{itemize}

\section{Related Work}
%
\subsection{Depth Reprojection-Based Video Generation.}
This line of work~\cite{zhang2025recapture,ren2025gen3c, mark2025trajectorycrafter, muller2024multidiff} estimates per-frame depth from an input and reprojects the resulting 3D structure along a target camera trajectory. 
The newly exposed or unprojected regions are subsequently predicted or inpainted to produce the final rendered video.
For instance,
GEN3C~\cite{ren2025gen3c} proposes a method that builds a spatio-temporal 3D cache from videos via depth estimation and unprojection, renders this cache along the user-specified camera trajectory, and conditions a video diffusion model on these renderings to achieve precise camera control and temporal consistency.
Meanwhile, TrajectoryCrafter~\cite{mark2025trajectorycrafter} introduces a dual-stream conditional video diffusion framework that takes as input both the source video and reprojection results, 
which are subsequently processed by a video inpainting model fine-tuned on a dataset generated using a double reprojection scheme.
Since this line of work relies on external depth estimation, 
its performance is ultimately limited by the accuracy of the estimated depth, 
which directly affects the magnitude of reprojection errors.

\subsection{Trajectory-Conditioned Video Generation.}
This line of work~\cite{van2024generative, bai2024syncammaster, bai2025recammaster, liang2025wonderland, voleti2024sv3d, sun2024dimensionx, xu2024camco} trains a video generation model that is explicitly conditioned on camera trajectories, using paired datasets of trajectories and corresponding videos. 
SynCamMaster~\cite{bai2024syncammaster} builds upon a pre-trained text-to-video model with a multi-view synchronization module to produce synchronized videos from diverse camera viewpoints. 
Trained with a hybrid data scheme and the released SynCamVideo dataset, it demonstrates open-domain novel-view video generation for stationary cameras.
ReCamMaster~\cite{bai2025recammaster} further leverages the generative capabilities of pre-trained text-to-video models through frame-dimension conditioning, and is trained with the MultiCamVideo dataset, a large-scale multi-camera synchronized video dataset built with Unreal Engine 5.
While effective, such trajectory-conditioned video generation methods trained on trajectory–video pair data often inherit biases from 
the underlying training data distribution. 

\begin{figure*}[t]
    \centering
    \includegraphics[width=\textwidth]{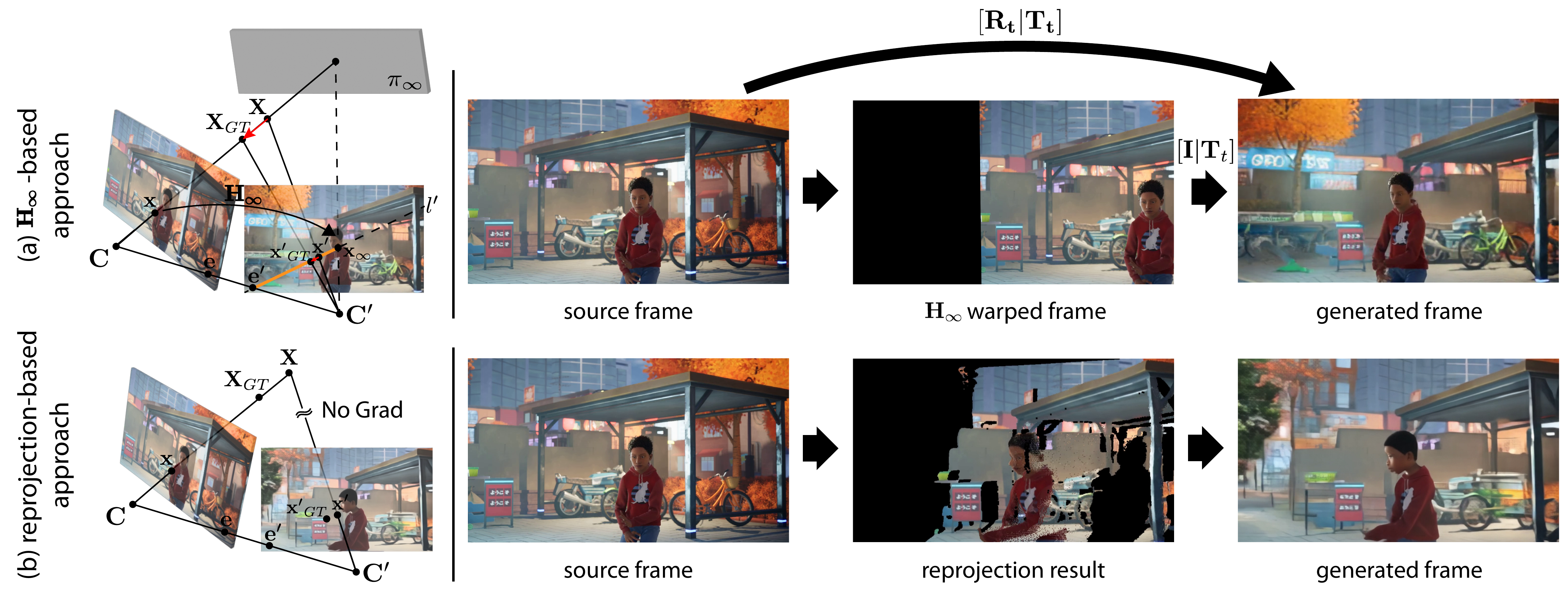}
    \vspace{-0.6cm}
    \caption{ 
        (a) \textbf{Infinite homography-based approach (ours).} By conditioning 
        on images warped by $\mathbf{H}{\infty}$, the model focuses on learning the parallax relative to the plane at infinity. This parallax is confined to the region between the epipole $\mathbf{e}'$ and $\mathbf{x}_{\infty}$, as visualized by the yellow segment on the epipolar line $l'$. 
        This spatial constraint helps the model to achieve 
        higher camera pose fidelity with reduced search space. 
        End-to-end training enables the network to implicitly refine the 3D geometry, correcting inaccuracies in $\mathbf{X}$. 
        (b) \textbf{Reprojection-based approach.} Inaccuracies in depth estimation lead to unreliable conditions, causing artifacts in the generated image. Since no gradients flow through the depth estimation network, the incorrect reprojection position $\mathbf{x'}$ remains fixed during training, hindering error correction.
    }
    \vspace{-0.2cm}
    \label{fig:motivation}
\end{figure*}

\section{Motivation}
~\cref{fig:motivation} illustrates the difference between the infinite homography-based approach and the reprojection-based approach at the frame level.
For a pair of cameras with centers $\mathbf{C}$ and $\mathbf{C'}$, 
the image from $\mathbf{C}$ is used to generate the corresponding view of camera $\mathbf{C'}$.

\subsection{Reprojection-based Approach.}
As illustrated in~\cref{fig:motivation} (b),
The reprojection-based approach first estimates the depth for a pixel $\mathbf{x}$ in camera $\mathbf{C}$, then reproject it to camera $\mathbf{C'}$, producing a corresponding position $\mathbf{x'}$ and yielding a reprojection image.
However, 
any inaccuracy in the predicted depth results in an unreliable condition being provided to the generative model, leading to artifacts in the synthesized image. Moreover, because no gradients flow through the depth estimation network, the inaccurate reprojection position $\mathbf{x'}$ remains fixed as a condition throughout training. 
Although the model can be trained to compensate for these errors, 
the challenge lies in handling noisy conditions, where it is unclear which parts are corrupted and to what extent, making the problem difficult to address. 

\subsection{Infinite Homography-based Approach. }
The infinite homography $\mathbf{H}_{\infty}$ represents the homography induced by the plane at infinity $\mathbf{\pi}_{\infty}$~\cite{hartley2003multiple}.
Given source and target camera intrinsic matrices $\mathbf{K}_{s}$ and $\mathbf{K}_{t}$, rotation matrix $\mathbf{R}$, translation vector $\mathbf{t}$, and normal $\mathbf{n}$ of a plane, the infinite homography $\mathbf{H}_{\infty}$ can be derived from the plane-induced homography $\mathbf{H}=\mathbf{K}_{t}(\mathbf{R}-\mathbf{t}\mathbf{n}^T/d)\mathbf{K}_{s}^{-1}$ by taking the limit as the distance $d$ to the plane approaches infinity:
\begin{equation}
    \mathbf{H}_{\infty} = \lim_{d\rightarrow\infty} \mathbf{H} = \mathbf{K}_{t}\mathbf{R}\mathbf{K}^{-1}_{s}.
    \label{eq:H_inf}
\end{equation}
For a pixel $\mathbf{x}$ with known depth $Z$ measured from the source camera, 
the reprojection $\mathbf{x'}$ to the target image is expressed as:
\begin{equation}
    \mathbf{x'}=\mathbf{K}_{t}\mathbf{R}\mathbf{K}^{-1}_{s}\mathbf{x} + \mathbf{K}_{t}\mathbf{t}/Z = \mathbf{H}_{\infty}\mathbf{x} + \mathbf{K}_{t}\mathbf{t}/Z
    \label{eq:projection}
\end{equation}
Notably, $\mathbf{H}_{\infty}$ does not depend on the translation or depth, 
enabling correspondence of image points at arbitrary depths when the camera undergoes pure rotation ($\mathbf{t}=\mathbf{0}$).
The term $\mathbf{K}_{t}\mathbf{t}/Z$ represents the parallax relative to the plane at infinity.
As shown in ~\cref{fig:motivation}(a), this parallax is restricted to the region between the epipole $\mathbf{e}'$ and $\mathbf{x}_{\infty}$, visualized by the yellow segment on the epipolar line $l'$.

Assuming that $\mathbf{K}_{s}$, $\mathbf{K}_{t}$, $\mathbf{R}$, and $\mathbf{t}$ are known, 
conditioning the network on the image warped by $\mathbf{H}_{\infty}$ 
allows the network to refer to a noise-free image under rotational transformation,
enabling it to focus on learning residual information, 
including $\mathbf{K}_{t}\mathbf{t}$ and the implicitly predicted depth $Z$, which modulates the final novel views.
Since parallax variations are constrained between the epipole $\mathbf{e}'$ and $\mathbf{x}_{\infty}$, 
this provides the model with strict boundary conditions (\cref{fig:motivation} (a) orange line).
This spatial constraint helps the model to improve camera-pose fidelity by effectively narrowing the search space.
Furthermore, unlike the reprojection-based approach, our method is trained in an end-to-end manner, enabling the network to implicitly refine the predicted depth $Z$ toward the ground truth as it learns to produce accurate parallax.
For computational efficiency, we warp the latent feature rather than the image itself during training.

\begin{figure*}[t]
    \centering
    \includegraphics[width=1.0\textwidth]{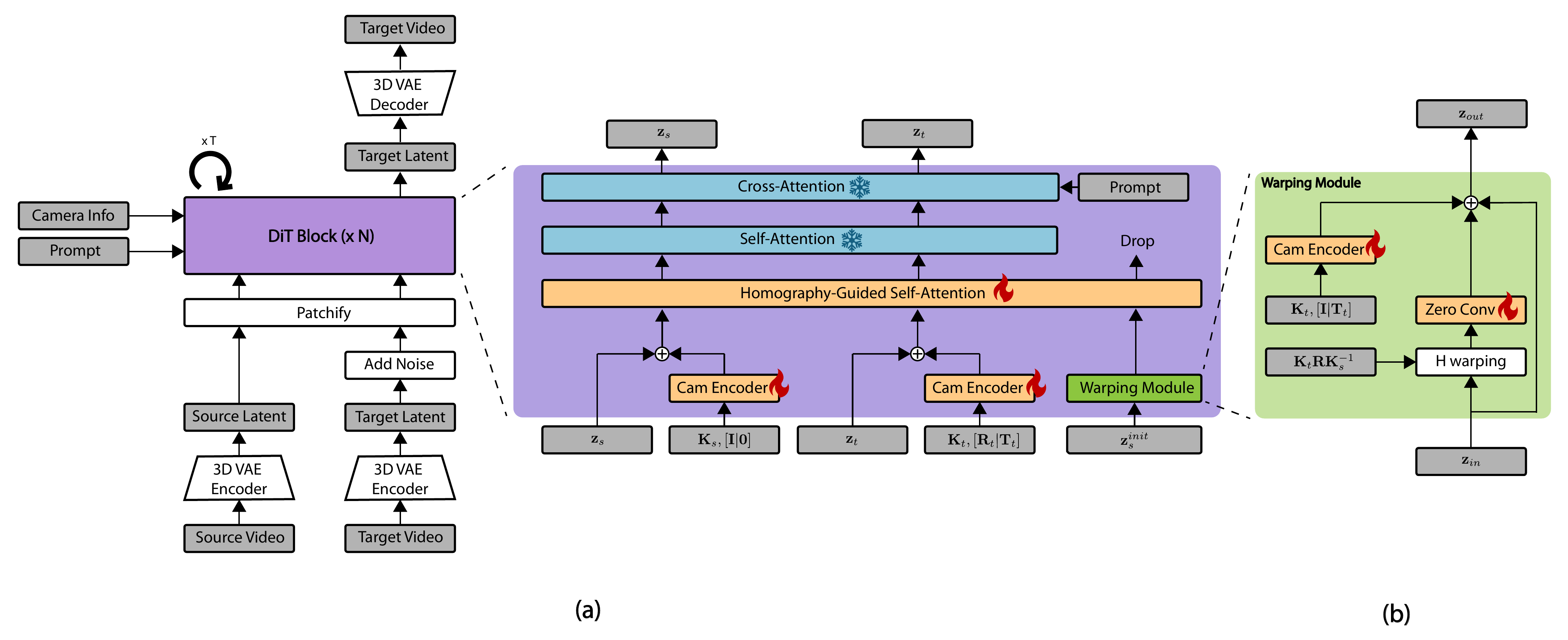}
\vspace{-0.8cm}
\caption{
    \textbf{Model Architecture Overview.} Our model builds upon Wan2.1, 
    training only newly introduced parameters while freezing pretrained weights. 
    (a) \textbf{DiT block with homography-guided self-attention layer.} 
    Homography-guided self-attention layer takes
    source, target, and warped latents combined with camera embeddings as input,
    and performs per-frame attention, ensuring temporal alignment.     
    By conditioning on warped latents, 
    the model enables rotation-aware reasoning and constrained parallax estimation. 
    Only source and target latents proceed to the subsequent Wan2.1 layers.  
    (b) \textbf{Warping module}. 
    This module warps the input latent with infinite homography 
    to handle rotation, then add camera embeddings for translation. 
    This decomposition simplifies reprojection to parallax estimation relative to plane at infinity, enabling higher camera trajectory fidelity.
    }
    \vspace{-0.3cm}
    \label{fig:model}
\end{figure*}

\section{Method}
Our goal is to perform novel-view video generation that faithfully follows a given target camera trajectory.
Specifically, given a source video $\mathbf{V}_{s}\in \mathbb{R}^{F \times C \times H \times W}$, 
source and target camera intrinsics $\mathbf{K}_s, \mathbf{K}_t \in \mathbb{R}^{3 \times 3}$, and 
target camera trajectory $\mathbf{T} \in \mathbb{R}^{F \times 3 \times 4}$ 
expressed relative to the source video's initial camera pose, 
our ~\modelname{} generates novel view video $\mathbf{V}_{t}\in \mathbb{R}^{F \times C \times H \times W}$ that faithfully follows the target camera trajectory and target intrinsic configuration.
Target camera trajectory is defined in a special Euclidean space $(\mathbf{R}, \mathbf{t}) \in SE(3)$, having rotation $\mathbf{R}_{t} \in \mathbb{R}^{3 \times 3}$ and translation $\mathbf{t}_{t} \in \mathbb{R}^3$.

In ~\cref{sec:model}, we present our model design tailored for the novel-view video synthesis task.
In ~\cref{sec:aug}, we describe our data augmentation strategy that enhances existing synthetic data for novel-view video synthesis across unconstrained camera trajectories and varying intrinsics.

\subsection{Model Architecture}
\label{sec:model}
The overall architecture of our model is illustrated in ~\cref{fig:model}.
We adopt the text-to-video model Wan2.1~\cite{wan2025} as our base architecture. 
To incorporate camera controllability while preserving Wan2.1's video generation capabilities trained on extensive datasets, 
we freeze the pretrained weights of Wan2.1 and train only the newly introduced camera encoder and the homography-guided self-attention layers. 
The new attention layers are initialized using weights from the corresponding pretrained transformer blocks.
We employ a camera encoder consisting of a linear layer with 16-dimensional input (comprising a flattened $3 \times 3$ rotation matrix, three translation parameters, focal lengths $f_x$, $f_y$, and principal points $c_x$, $c_y$) and $d$-dimensional output to encode camera poses, where $d$ represents the hidden dimension of the self-attention. 
For the latent representation $\mathbf{z} \in \mathbb{R}^{b \times f \times h \times w \times d}$, 
we add the corresponding camera embedding $e_c \in \mathbb{R}^{b \times f \times d}$, broadcasting it along the spatial dimension. 
To ensure consistent camera conditioning, we share the camera encoder across each DiT block for different camera encodings.

\subsubsection{Warping Module.} 
Our warping module is illustrated in~\cref{fig:model} (b).
The warping module, motivated by~\cref{eq:projection}, utilizes an infinite homography to warp the latent representation.
Since target camera poses are defined relative to the source video's first frame, 
the warping module warps the source latent of the initial frame ($\mathbf{z}^{\text{init}}_s$) using the homography derived from the camera rotation and intrinsics, 
as described in~\cref{eq:H_inf}. (Warping can also be extended to all frames of the source video, i.e., per-frame warping, 
see Supplementary~\cref{sup:perframe} for a comparison.)
The warped result is then added to the original $\mathbf{z}^{\text{init}}_s$ through a convolutional layer initialized to zero, functioning as a residual connection. 
Next, camera embeddings, which encode the target's translation and intrinsics, are incorporated to reflect the second term in~\cref{eq:projection}. 
This design simplifies the reprojection estimation under target camera poses by reducing it to parallax relative to $\pi_{\infty}$ estimation, helping the model 
achieve higher camera-pose fidelity.
The effectiveness of the warping module is validated through ablation studies presented in~\Cref{tab:quan_ablation}.

\subsubsection{Homography-Guided Attention Layer.} 
The DiT block incorporating the homography-guided attention layer is illustrated in~\cref{fig:model}(a).
For frame index $i$, we process three types of latents: 
the source latent $\mathbf{z}_{s}^{i}$, target latent $\mathbf{z}_{t}^{i}$, and warped latent $\mathbf{z}_{w}^{i}$. 
Each latent is combined with its corresponding camera embedding before being fed into the attention layer.
Specifically, to obtain the camera embedding for the target latent, we concatenate the user-specified target intrinsics $\mathbf{K}_{t}$, rotation $\mathbf{R}_{t}$, and translation $\mathbf{t}_{t}$, and feed it into the camera encoder. 
For the source latent,  as we assume that the source camera poses are unknown, we form a camera input vector by concatenating the source intrinsics with an identity pose $[\mathbf{I}|\mathbf{0}]$. 
This vector is then replicated across frames and encoded using the camera encoder.
The homography-guided attention layer performs per-frame attention by spatially concatenating these three input latents.
The resulting concatenated latents have shape $\mathbf{z}_{c}\in\mathbb{R}^{bf \times 3hw \times d}$, 
where frames are processed as individual batch items within the attention mechanism.
This structure ensures temporal alignment by allowing the target frames to reference corresponding source frames at the same timestamps. 
Additionally, the concatenated warped latents help the model better reason about rotation-induced view transformations,
which in turn leads to 
constrained parallax estimation within static regions.
After passing through the homography-guided attention layer, the concatenated features are split and reshaped into three tensors of shape 
$\mathbb{R}^{b \times fhw \times d}$.  
Finally, we discard the warped latent $\mathbf{z}^{w}_{i}$ and feed only the source and target latents into Wan2.1's self-attention layer.
During processing through the pretrained Wan2.1 layers, the paired source and target latents are treated as a unified batch.
For a more detailed model architecture, please refer to our Supplementary~\cref{sup_sec:model}.

\begin{figure}[t]
    \centering
    \includegraphics[width=\columnwidth]{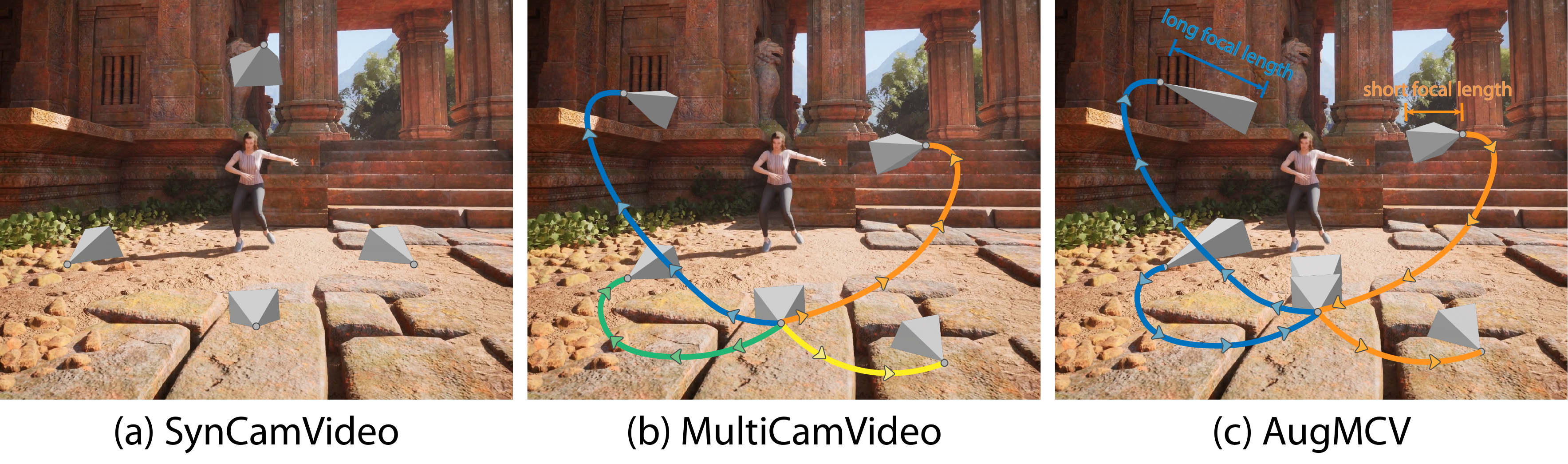}
    \caption{ 
        Visualization of synchronized multi-view synthetic video datasets. 
        Different trajectories are visualized in different colors. 
        (a) \textbf{SynCamVideo.} Captured with stationary cameras placed at distinct positions.
        (b) \textbf{MultiCamVideo.} Captured with dynamic cameras following diverse trajectories, all sharing the same initial frame.
        (c) \textbf{AugMCV.} An augmented version of MultiCamVideo with varied starting poses and different focal lengths.
    }
    \vspace{-0.3cm}
    \label{fig:dataset_ours}
    \vspace{-0.3cm}
\end{figure}

\subsection{Data Preparation}
In this section, we describe our data augmentation strategy that enhances existing synthetic data for novel-view video synthesis with unconstrained camera trajectories and varying intrinsics. We utilize the MultiCamVideo Dataset~\cite{bai2025recammaster} for augmentation and will refer to the augmented version as ~\dataname{} (Augmented MultiCamVideo) Dataset for brevity.
\label{sec:aug}
\subsubsection{Trajectory Augmentation}
\label{sec:aug_ext} 
The MultiCamVideo dataset provides 10 trajectories and their corresponding videos for each scene. 
These trajectories cover a diverse set of camera motions, including pan, tilt, translation, arc, random, and static movements.
A notable feature of this data is that all 10 cameras within each scene share identical starting positions, ensuring consistent initialization across viewpoints. 
Consequently, when two videos from the same scene are randomly sampled as the source and target videos for training, their first frames are always identical. 
We empirically observe that models trained on this dataset exhibit a bias toward reproducing the source video’s first frame, even when conditioned to generate it from a different viewpoint.

Building on these properties, we introduce a data augmentation approach 
to eliminate such bias from initial frame conditioning. 
Our key observation is that while all trajectories share the same starting frame, their 
remaining frames diverge 
across different videos. 
Leveraging this property, we randomly sample two distinct videos from the same scene and construct an augmented sequence by reversing the first video and concatenating it with the second one. 
Since the final frame of the reversed video coincides with the initial frame of the second video, we remove the redundant first frame of the latter, resulting in an augmented video of 161 frames (81 + 80). 
To align with the pre-training strategy of the Wan2.1 base model, we segment 81 frames from videos within the same scene, ensuring that they share the same starting timestamps for synchronization. 
~\cref{fig:dataset_ours} illustrates how the dataset configuration changes under our trajectory augmentation approach.
This approach preserves temporal alignment while introducing variability in the initial frame, 
enabling the model to learn more generalizable trajectory representations and mitigating dataset-specific biases.

\subsubsection{Intrinsic Augmentation}
\label{sec:aug_int}
The MultiCamVideo Dataset incorporates four distinct focal lengths: 18mm, 24mm, 35mm, and 50mm, with the focal length remaining constant within each scene.
Each focal length configuration contains 3,400 scenes, summing up to 13,600 scenes in total.
While the dataset inherently encompasses multiple intrinsic parameters, our empirical analysis reveals that models trained on this dataset demonstrate a bias toward generating videos that preserve the focal length of the source video.
This limitation arises from the training paradigm wherein both source and target videos are sampled from identical scenes with fixed focal length.
Such focal length consistency prevents the model from learning the underlying relationship between focal length variations and their corresponding effects on video generation.
To alleviate this bias, we incorporate intrinsic augmentation.

Specifically, given the trajectory-augmented scene with focal length $f_{scene}$, we randomly sample the new focal length $f_{\text{new}} \in \{x\in\{18\text{mm}, 24\text{mm}, 35\text{mm}, 50\text{mm}\} | x > f_{scene}\}$, and apply intrinsic augmentation process to the input video. 
Intrinsic augmentation is performed by resizing the video based on the ratio of source and target focal lengths, followed by center cropping.
The detailed process is provided in the Supplementary~\cref{sup_sec:aug_focal}.

\subsubsection{Video Pair Selection}
\label{sec:aug_select}
To train the video-to-video model for the purpose of novel-view video generation, paired video data is required. 
In this setup, the source video represents the user input, while the target video serves as the ground truth, 
guiding the model’s output to generate a video that follows the specified trajectory. 
To create these video pairs,  
we randomly sample two cameras out of the ten synchronized ones from an identical scene, using one as the source video and the other as the target video.
For each video, we apply focal length augmentation with a probability of 0.5. 
Although we only augment focal lengths in the ascending direction, using augmented videos as either source or target ensures that the selected video pairs encompass both focal length increase and decrease scenarios.
This video pair configuration strategy allows for coverage of arbitrary trajectory patterns, as well as the generation of narrow and wide field-of-view videos.

\section{Experiments}
\subsection{Experiment Settings}
\subsubsection{Implementation Details.}
We use the pretrained Wan2.1 model as our backbone.
Since Wan2.1 requires a text description of the generated video, 
we employ LLaVA~\cite{liu2024llavanext} to extract descriptive text from the source video, which we then use as the text input.
For training, we use the train split of the \dataname{} dataset introduced in~\cref{sec:aug}, which consists of 47,432 scenes, each containing 10 trajectories.
For quantitative evaluation, we train on videos of length 81-frame videos at resolution of \(480\times832\) for 15k iterations using four H100 GPUs (batch size 8), taking approximately one week.
For ablations, we train on 41-frame videos at \(320\times544\) resolution for 20k iterations using four H100 GPUs (batch size 32), which takes about four days, to ensure efficiency and ease of experimentation. 
For both resolutions, we use the Adam optimizer with a weight decay of 0.01 and a learning rate of 1e-5.

\subsubsection{Baselines.}
We compare our method against state-of-the-art camera-controlled video generation algorithms, including ReCamMaster~\cite{bai2025recammaster}, TrajectoryCrafter~\cite{mark2025trajectorycrafter}, and GEN3C~\cite{ren2025gen3c}.
ReCamMaster is a trajectory-conditioned video generation approach trained on synthetic trajectory–video pair datasets.
In contrast, TrajectoryCrafter and GEN3C are reprojection-based approaches that estimate depth from the source video and condition the generation process on the reprojection results.

\subsubsection{Evaluation.}
Our method and baselines are evaluated on two datasets.

\datasetheading{(1) \dataname{} Dataset.} 
Experiments are conducted on the test split of ~\dataname{} dataset, which consists of 168 scenes.
Each scene contains one video captured with a static camera and ten videos recorded along dynamic trajectories.
Using the static-camera clip as a source, we generate one video for each target trajectory, resulting in a total of 1,680 generated videos.
Among the 168 test scenes, 96 scenes have source and target videos with identical focal lengths (referred to as \textit{shared intrinsics}), while the remaining 72 scenes involve different focal lengths between source and target videos (referred to as \textit{different intrinsics}).
The generated videos are evaluated against their corresponding ground-truth videos using PSNR, SSIM, and LPIPS~\cite{zhang2018unreasonable}.

\datasetheading{(2) WebVid Dataset.}
A random subset of 100 source videos is selected from the WebVid~\cite{webvid_bain21} dataset to evaluate performance in real-world scenarios. 
For each source video, we generate 20 synthetic videos using different camera trajectories. Specifically, 10 trajectories share the same initial camera pose as the first frame of the source video, while the remaining 10 trajectories start from different initial camera poses.
This setup yields a total of 2,000 generated synthetic videos.
As the dataset does not provide camera intrinsics for the source videos, we estimate the source intrinsics at inference time using UniDepth~\cite{piccinelli2024unidepth}.
Since no ground-truth videos are available for the target trajectories, we evaluate perceptual fidelity using FID~\cite{heusel2017gans} for frame-level quality and FVD~\cite{unterthiner2019fvd} for video-level quality.
To assess camera pose fidelity, we measure rotation and translation errors~\cite{he2024cameractrl} between the target trajectories and those estimated from the generated videos using ViPE~\cite{huang2025vipe}.
To further evaluate the preservation of subject appearance and temporal continuity, we incorporate Subject Consistency (S.C.) and Motion Smoothness (M.S.) metrics from VBench~\cite{huang2024vbench}.
%

\begin{figure*}[t]
    \centering
    \includegraphics[width=\textwidth]{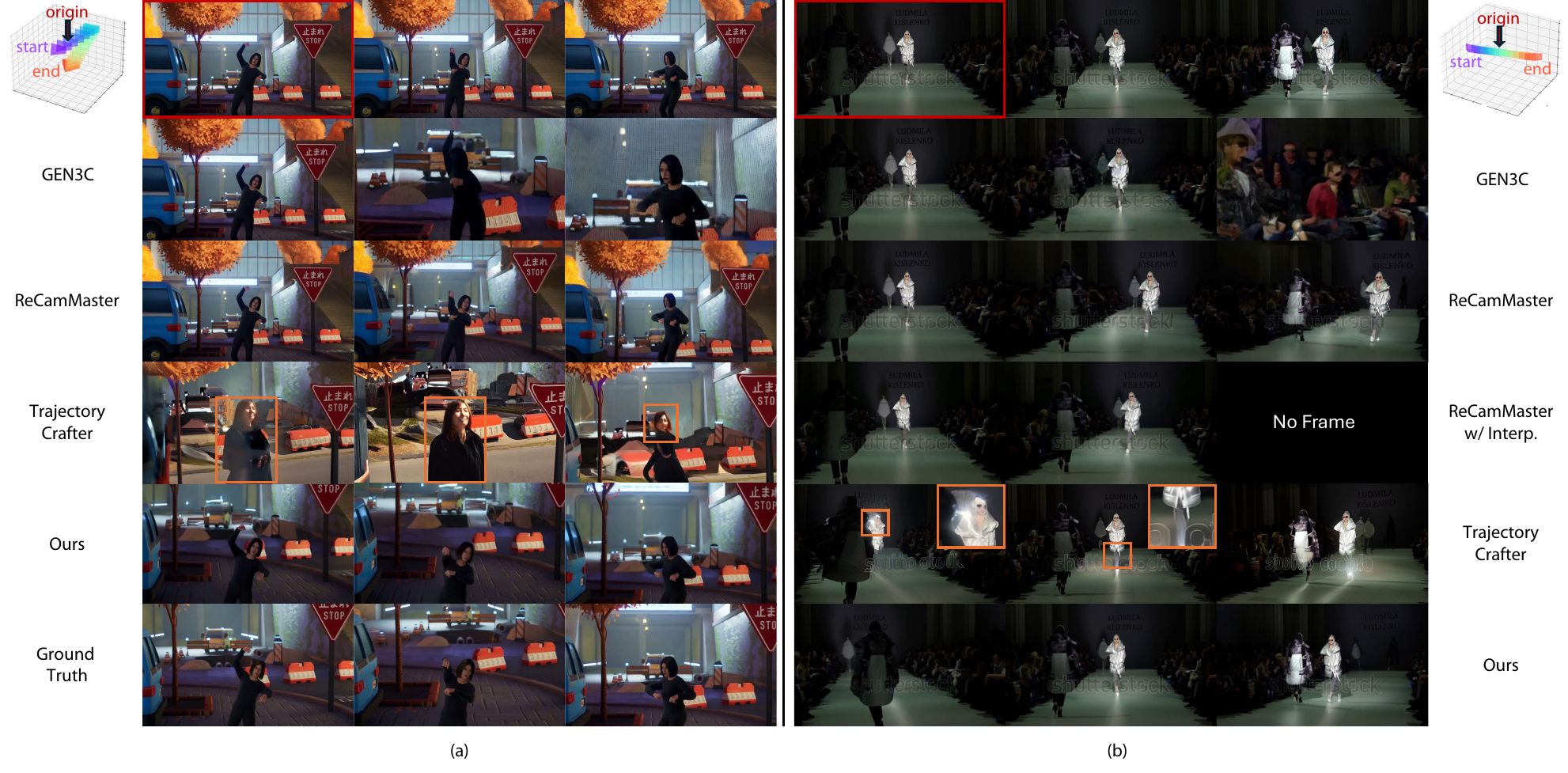}
    \vspace{-0.5cm}
    \caption{
        \textbf{Qualitative Comparison.} (a) shows results on the test split of the \dataname{} dataset, and (b) presents results on the WebVid dataset. In both cases, GEN3C and ReCamMaster fail to perform proper viewpoint transformations, largely preserving the initial frame of the source video. ReCamMaster further fails to reflect pose changes in the initial frame even when trajectory interpolation is applied (ReCamMaster \textit{w/ Interp}). TrajectoryCrafter introduces artifacts due to inaccurate reprojection (highlighted in the orange box). In contrast, our method achieves natural pose transitions while maintaining high visual quality throughout the sequence. Best viewed in zoom. }
        \vspace{-0.3cm}
    \label{fig:qual_comparison}
\end{figure*}

\begin{table}[t]
\centering
\small 
\caption{Quantitative comparison on the \dataname{} test set. The best and second-best results are \textbf{bold} and \underline{underlined}, respectively.}
\label{tab:quan_indomain}

\newcolumntype{Y}{>{\centering\arraybackslash}X}
\begin{tabularx}{\columnwidth}{l YYY YYY}
\toprule
& \multicolumn{3}{c}{Shared Intrinsics} & \multicolumn{3}{c}{Different Intrinsics} \\
\cmidrule(lr){2-4} \cmidrule(lr){5-7} 
Method & PSNR$\uparrow$ & SSIM$\uparrow$ & LPIPS$\downarrow$ & PSNR$\uparrow$ & SSIM$\uparrow$ & LPIPS$\downarrow$ \\
\midrule
GEN3C~\cite{ren2025gen3c} & 16.891 & 0.479 & 0.548 & 17.449 & 0.525 & 0.467 \\
ReCamMaster~\cite{bai2025recammaster} & 21.130 & 0.617 & 0.420 & \underline{19.665} & 0.558 & 0.510 \\
TrajectoryCrafter~\cite{mark2025trajectorycrafter} & \underline{21.228} & \underline{0.660} & \underline{0.296} & 19.557 & \underline{0.586} & \underline{0.390} \\
\rowcolor{gray!15} 
\textbf{Ours} & \textbf{22.677} & \textbf{0.718} & \textbf{0.246} & \textbf{22.261} & \textbf{0.699} & \textbf{0.270} \\
\bottomrule
\end{tabularx}
\end{table}

\begin{table}[h]
\centering
\caption{Quantitative comparison on the WebVid dataset. S.C. and M.S. denote Subject Consistency and Motion Smoothness. The best and second-best results are \textbf{bold} and \underline{underlined}, respectively.}
\vspace{-0.2cm}
\label{tab:quan_outdomain}

\newcolumntype{Y}{>{\centering\arraybackslash}X}

\begin{tabularx}{\columnwidth}{l YYYYYY}
\toprule
Method & RotErr$\downarrow$ & TransErr$\downarrow$ & FID$\downarrow$ & FVD$\downarrow$ & S.C.$\uparrow$ & M.S.$\uparrow$ \\
\midrule
GEN3C~\cite{ren2025gen3c} & 9.588 & 3.012 & 43.790 & 331.768 & 0.908 & 0.990 \\
ReCamMaster~\cite{bai2025recammaster}            & 8.375 & 1.027 & 39.930 & 302.025 & 0.944 & \textbf{0.994}  \\
ReCamMaster w/ interp. & 9.016 & 1.590 & 35.720 & 303.012 & \underline{0.946} & \underline{0.993} \\
TrajectoryCrafter~\cite{mark2025trajectorycrafter} & \underline{5.007} & \underline{0.735} & \underline{30.877} & \underline{289.879} & 0.940 & 0.991\\
\rowcolor{gray!15} 
Ours                   & \textbf{3.162} & \textbf{0.438} & \textbf{29.702} & \textbf{286.952} & \textbf{0.949} & \underline{0.993}\\
\bottomrule
\end{tabularx}
\end{table}

\subsection{Qualitative Results}\label{sec:exp_qual}
\cref{fig:qual_comparison} (a) and (b) show qualitative results on the AugMCV test split and the WebVid dataset, respectively. 
In both cases, the first row shows the source video, followed by the generated results from each method. For the \dataname{} dataset, the last row displays the ground-truth target video, whereas for the WebVid dataset, only the generated results are available since ground-truth target views are not provided. Frames are uniformly sampled along the camera trajectory for visualization.

Across both datasets, ReCamMaster consistently preserves the initial frame of the source video, which we attribute to a bias in its training data, where all paired source-target trajectories originate from the same initial frame of the source video. 
To mitigate this bias, we applied a frame interpolation strategy and visualized the results in the fourth row of \cref{fig:qual_comparison}(b). 
Specifically, eight identical frames were prepended to the source video, and their corresponding camera poses were linearly interpolated between the identity pose and the target pose of the first frame. 
The first eight frames were then removed from the generated video before evaluation to exclude the interpolated portion. 
Although this strategy enables ReCamMaster to synthesize different viewpoints at the initial frame, the results of \textit{ReCamMaster w/ interp.} in \cref{fig:qual_comparison}(b) still exhibit alignment errors, indicating that interpolation alone cannot fully address its trajectory bias.

TrajectoryCrafter, a reprojection-based method, successfully performs viewpoint transformation but fails to maintain the source appearance throughout the sequence due to reprojection errors introduced by inaccurate depth estimation. 
GEN3C also relies on reprojection, but fails to transform the viewpoint of the initial frame.
We conjecture that this result stems from a bias in its image-to-video backbone toward preserving the initial source frame. 
This observation is consistent with that reported in GCD~\cite{van2024generative}.

In contrast, our method, empowered by the infinite-homography-based warping module and trajectory-intrinsic augmentation, achieves consistent viewpoint alignment and maintains high visual coherence throughout the trajectory. 

\begin{figure}[t]
    \centering
    \includegraphics[width=\columnwidth]{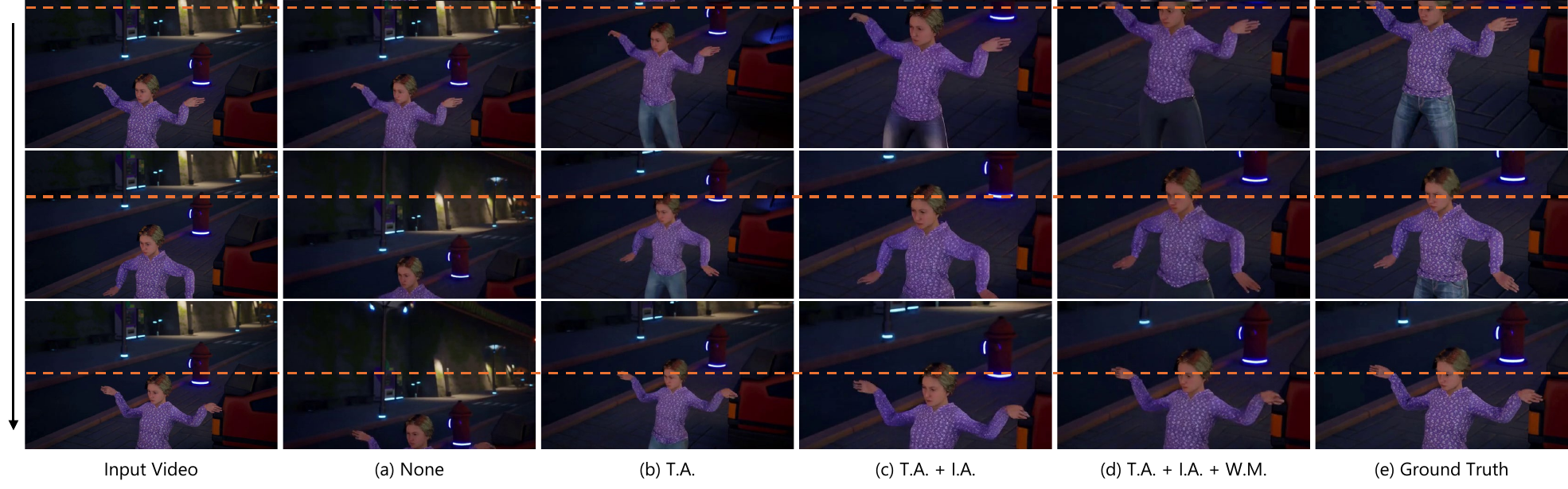}
    \caption{ 
        Qualitative ablation study of proposed components. 
        From left to right, we incrementally integrate: (b) Trajectory Augmentation (T.A.), (c) Intrinsic Augmentation (I.A.), and (d) our full model with the Warping Module (W.M.). 
        To evaluate camera pose fidelity, a dashed orange guideline is aligned with the eye level in the Ground Truth (e). While (a) None, (b) T.A., and (c) T.A.+I.A. exhibit significant vertical drift and misalignment relative to the guide line, the full configuration (d) maintains consistent alignment, demonstrating that all three components are essential for accurate video generation.
    }
    \label{fig:qual_ablation}
\end{figure}

\subsection{Quantitative Results}
\Cref{tab:quan_indomain} presents quantitative evaluation results on~\dataname{} dataset for two scenarios: 
(1) source and target videos with identical camera intrinsics, and
(2) source and target videos with different camera intrinsics.
Across all scenarios, our method consistently outperforms the baselines in all metrics, demonstrating that the videos generated by our method are clearly closer to the ground truth.

\Cref{tab:quan_outdomain} presents the quantitative evaluation results on the WebVid dataset. Our method consistently outperforms the baseline approaches in terms of camera pose accuracy(RotErr, TransErr), visual fidelity(FID, FVD), and temporal consistency (Subject Consistency, Motion Smoothness). In particular, our approach demonstrates a significant improvement in camera pose accuracy over the baselines.
Depth-based approaches, such as GEN3C and TrajectoryCrafter, exhibit lower S.C. scores due to shape morphing caused by depth reprojection errors, whereas \modelname{} effectively preserves subject integrity. 
While ReCamMaster achieves the highest M.S., its substantial pose error suggests that it prioritizes general temporal smoothness at the expense of precise trajectory adherence.
We provide an extended comparison with ReCamMaster in the supplementary~(\cref{sup_sec:rcm_compare}).
Ultimately, our method achieves the most harmonious balance across all metrics, ensuring both precise camera control and superior perceptual quality.

\begin{table}[t]
\centering
\small 
\caption{Ablation study conducted on the \dataname{} test set. The best and second-best results are \textbf{bold} and \underline{underlined}, respectively.}
\label{tab:quan_ablation}

\newcolumntype{Y}{>{\centering\arraybackslash}X}

\begin{tabularx}{\columnwidth}{YYY YYY YYY}
\toprule
\multicolumn{3}{c}{Components} & \multicolumn{3}{c}{Shared Intrinsic} & \multicolumn{3}{c}{Different Intrinsic} \\
\cmidrule(lr){1-3} \cmidrule(lr){4-6} \cmidrule(lr){7-9} 
Aug.Traj. & Aug.Intr. & Warp & PSNR$\uparrow$ & SSIM$\uparrow$ & LPIPS$\downarrow$ & PSNR$\uparrow$ & SSIM$\uparrow$ & LPIPS$\downarrow$ \\
\midrule
$\square$ & $\square$ & $\square$ & 19.228 & 0.562 & 0.427 & 18.480 & 0.507 & 0.525 \\
$\checkmark$ & $\square$ & $\square$ & 20.820 & 0.615 & 0.353 & 18.865 & 0.523 & 0.499 \\
$\checkmark$ & $\checkmark$ & $\square$ & \underline{22.807} & \underline{0.680} & \underline{0.250} & \underline{21.866} & \underline{0.649} & \underline{0.293} \\
\addlinespace[0.5ex] 
\rowcolor{gray!15} 
$\checkmark$ & $\checkmark$ & $\checkmark$ & \textbf{24.412} & \textbf{0.733} & \textbf{0.198} & \textbf{24.311} & \textbf{0.720} & \textbf{0.203} \\
\bottomrule
\end{tabularx}
\end{table}

\subsection{Ablation Study}
\subsubsection{Analysis of Proposed Components. }
We perform an ablation study on ~\dataname{} test set by progressively adding each of the proposed components.
As shown in~\Cref{tab:quan_ablation} and~\cref{fig:qual_ablation}, the baseline without warping module and data augmentation fails to capture both the target trajectory and intrinsics, yielding the lowest scores in all settings.
Adding trajectory augmentation improves all scores in~\cref{tab:quan_ablation} and results in a movement consistent with the target trajectory in~\cref{fig:qual_ablation}; 
however, it still fails to reflect the target intrinsics, offering comparable performance to the baseline under \textit{different intrinsics} setting in~\cref{tab:quan_ablation}. 
Combining intrinsic augmentation with trajectory augmentation enables the model to better understand both shared and different intrinsic settings, 
although a performance gap relative to the ground truth persists. 
Finally, introducing the warping module, which explicitly warps the source latent using the target pose and intrinsics, proves crucial for accurately aligning the results with the ground truth, leading to substantial qualitative and quantitative improvements.
We further compare warping in the latent space against warping in pixel space in the supplementary~(\cref{sup_sec:warp_compare}).

\begin{table}[t]
\centering
\caption{Ablation study of our augmentation strategies on WebVid dataset. Best scores per metric are in \textbf{bold}.}
\vspace{-0.2cm}
\label{tab:quan_ablation2}

\newcolumntype{Y}{>{\centering\arraybackslash}X}

\begin{tabularx}{\columnwidth}{l YYYYYY}

\toprule
Method & Rot.$\downarrow$ & Trans.$\downarrow$ & FID$\downarrow$ & FVD$\downarrow$ & S.C.$\uparrow$ & M.S.$\uparrow$ \\
\midrule
\textit{w}/ MCV+SCV & 4.158 & 1.441 & 47.949 & 235.958 & 0.950 & \textbf{0.992} \\
\rowcolor{gray!25} 
\textit{w}/ AugMCV & \textbf{3.368} & \textbf{0.839} & \textbf{40.384} & \textbf{235.563} & \textbf{0.952} & \textbf{0.992}\\
\bottomrule
\end{tabularx}
\end{table}
\subsubsection{Effectiveness of our augmentation strategy.}
We evaluate the proposed trajectory-intrinsic augmentation scheme by comparing our model trained on the \dataname{} dataset with a model trained on the mixture of the MultiCamVideo (MCV) dataset and the SynCamVideo (SCV) dataset. 
The SynCamVideo dataset is a synchronized multi-camera video dataset rendered using Unreal Engine 5, where the cameras are stationary and their corresponding poses are provided.
Unlike the MultiCamVideo dataset, the SynCamVideo dataset has different initial frame poses for the source and target videos. 
Jointly training on both datasets is expected to mitigate the bias inherent in each dataset. 
However, as shown in \Cref{tab:quan_ablation2}, our model trained with~\dataname{} outperforms the joint training approach, highlighting the superior effectiveness of our trajectory-intrinsic augmentation. 

\section{Conclusion}
In this work, we introduced \modelname{}, a depth-free framework that achieves precise camera control in video generation through infinite homography warping. 
By encoding camera rotations in the latent space and learning residual parallax, our method removes the dependence on external depth estimation modules and erroneous reprojection results, resulting in high camera-pose accuracy without depth prior. 
Combined with our trajectory-intrinsic augmentation strategy, \modelname{} demonstrates superior performance across diverse camera motions and focal lengths.
InfCam consistently outperforms depth reprojection-based and trajectory-conditioned approaches across various metrics.
Moving forward, we aim to extend our framework to extreme novel view synthesis to overcome the current constraints detailed in Supplementary~\cref{sup_sec:limit}, particularly for unobserved scene regions.

\section*{Acknowledgements}
This work was supported by Institute for Information \& communications Technology Planning \& Evaluation(IITP) grant funded by the Korea government(MSIT) (RS-2019-II190075, Artificial Intelligence Graduate School Program(KAIST)).
This work was supported by Institute of Information \& Communication Technology Planning \& Evaluation (IITP) grant funded by the Korea government (MSIT) (No. RS-2025-02263841, Development of a Real-time Multimodal Framework for Comprehensive Deepfake Detection Incorporating Common Sense Error Analysis).
This work was supported by the National Supercomputing Center with supercomputing resources including technical support (KSC-2025-CRE-0188).
This research was supported by the ``Advanced GPU Utilization Support Program'' funded by the Government of the Republic of Korea (Ministry of Science and ICT).

%
%
\bibliographystyle{splncs04}
\bibliography{main}

\input{supple}

\end{document}

%% file: preamble.tex
\usepackage{eccv}

\usepackage{eccvabbrv}

\usepackage{graphicx}
\usepackage{booktabs}

\usepackage{graphicx}
\usepackage{cuted}
\usepackage{capt-of}

\usepackage{booktabs} 
\usepackage{multirow}
\usepackage{tabularx} 
\usepackage{makecell} 

\usepackage{algorithm}
\usepackage{algpseudocode}
\usepackage{enumitem}
\usepackage[table]{xcolor} 
\usepackage{wrapfig}

\newcommand{\modelname}{InfCam} 
\newcommand{\dataname}{AugMCV}

\newcommand{\datasetheading}[1]{%
  \vspace{0.5em}%
  \noindent{\textit{#1}}\enspace\ignorespaces
}

\newcommand*\samethanks[1][\value{footnote}]{\footnotemark[#1]}

\usepackage[accsupp]{axessibility}  

\usepackage{hyperref}

\usepackage{orcidlink}

%% file: supple.tex
\clearpage
\setcounter{page}{1}

\begin{figure*}[t]
    \centering
    \includegraphics[width=0.99\textwidth]{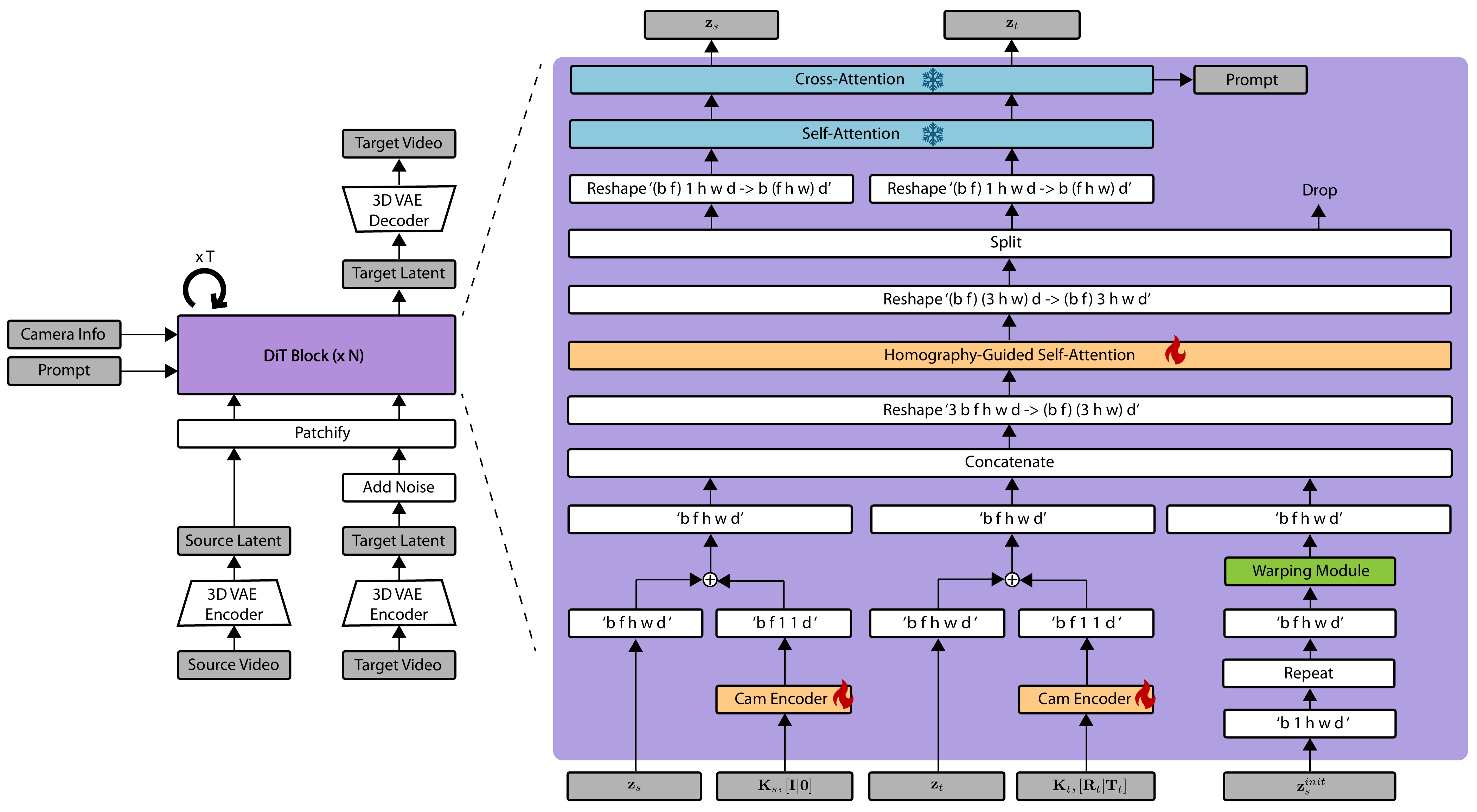}
    \captionof{figure}{
    This figure illustrates our model overview along with the dimensions of latent features. Here, $b$ denotes the batch size, $f$ indicates the latent frame count, $w$ and $h$ represent latent width and height, and $d$ represents the feature dimension. The dimensional notation follows the einops convention.}
    \label{fig:model_detail}
\end{figure*}
\section{Model Architecture Details}
\label{sup_sec:model}
~\cref{fig:model_detail} illustrates the overall architecture of the proposed ~\modelname{} with its dimensional specifications. 
As described in~\cref{sec:model}, the homography-guided self-attention layer performs per-frame attention by spatially concatenating three input latents: the source, target, and warped latents. The resulting concatenated latent $\mathbf{z}_{c}\in\mathbb{R}^{bf \times 3hw \times d}$ treats frames as individual batch items within the attention mechanism, where spatial features from all three latents are concatenated along the spatial dimension.

After passing through the homography-guided self-attention layer, $\mathbf{z}_{c}$ is split and reshaped back into three separate tensors, each with shape $\mathbb{R}^{b \times fhw \times d}$. Finally, we discard the warped latent and feed only the source and target latents into Wan2.1's self-attention layer. During processing through the pretrained Wan2.1 layers, the paired source and target latents are treated as a unified batch.

\section{Data Preparation Details}
\begin{figure*}[t]
    \centering
    \includegraphics[width=0.99\textwidth]{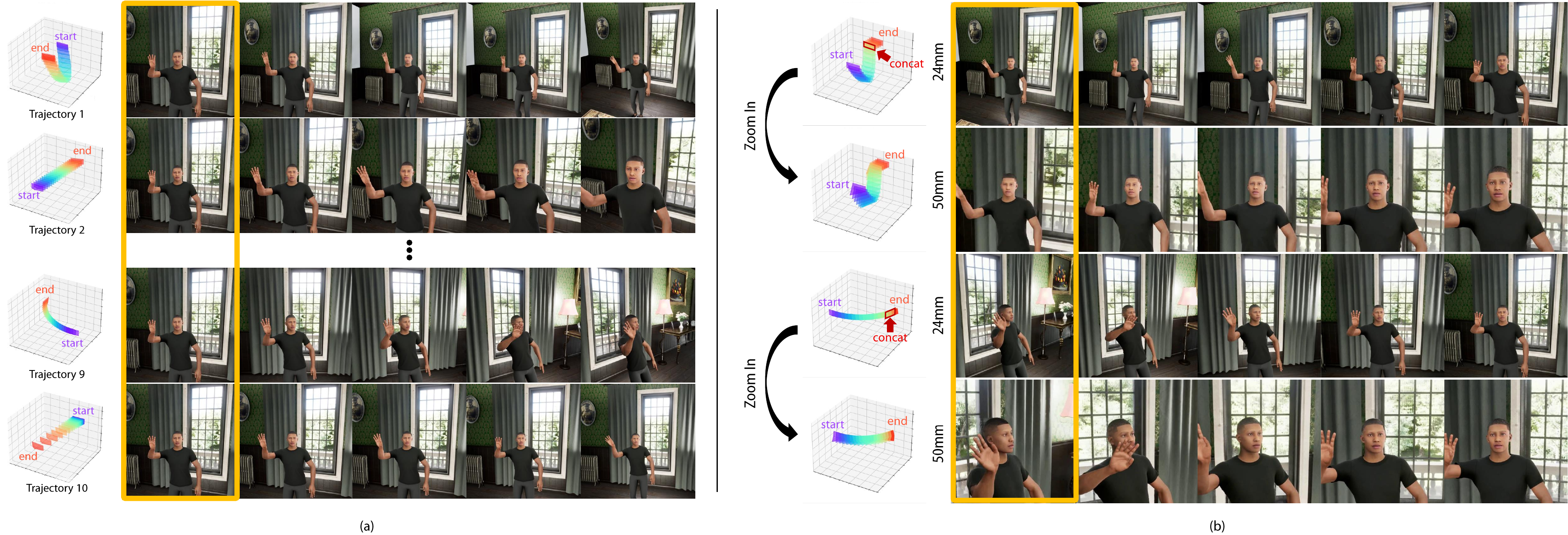}
    \caption{
        \textbf{(a)} Example scene from the MultiCamVideo dataset. 
        Each row shows a camera trajectory and its corresponding video, with all videos sharing the same initial frame (highlighted in yellow). 
        \textbf{(b)} Augmented dataset examples. 
        Rows 1\&3 show trajectory augmentation, which preserves temporal alignment among trajectories while introducing variation in the initial frame selection (highlighted in yellow). 
        Rows 2\&4 show focal-length augmentation, enabling the model to learn the relationship between focal-length changes and their visual effects in video generation.
    }
    \label{fig:dataset_before_after}
\end{figure*}

\subsection{Trajectory Augmentation. }
~\cref{fig:dataset_before_after} (a) shows an example scene from the MultiCamVideo dataset~\cite{bai2025recammaster}.
The MultiCamVideo dataset provides 10 trajectories and their corresponding videos for each scene. 
These trajectories cover a diverse set of camera motions, including pan, tilt, translation, arc, random, and static movements.
A notable property of this dataset is that all 10 cameras within each scene share an identical starting position (highlighted in yellow), which ensures that the first frame of the target video is always identical to the first frame of the source video.
We hypothesize that these dataset constraints arise from the fundamental requirement that camera viewpoints must maintain sufficient frustum overlap for the purpose of novel-view video generation, 
ensuring the task does not collapse into pure video generation.
To reduce this bias, we perform the data augmentation introduced in~\cref{sec:aug}.
~\cref{fig:dataset_before_after} (b) illustrates an augmented trajectory example and its corresponding focal-length (intrinsics) augmentation.
The trajectories illustrated in the first and third rows of~\cref{fig:dataset_before_after} (b) are constructed from the trajectory pairs (Trajectory 1, Trajectory 2) and (Trajectory 9, Trajectory 10) in~\cref{fig:dataset_before_after} (a) by reversing the first trajectory and concatenating it with the second one. 
From the resulting sequence, a segment of length 81 frames is randomly cropped to match the supported sequence length of the Wan2.1 base model.

\subsection{Intrinsic Augmentation. }\label{sup_sec:aug_focal}
Intrinsic augmentation is applied to the augmented trajectories. 
For illustration, the first and third rows in~\cref{fig:dataset_before_after} (b) depict the trajectories prior to intrinsic augmentation, 
whereas the second and fourth rows display their augmented counterparts.
~\cref{sup_alg:aug_focal} illustrates how we augment the focal length for a given scene. 
This process involves resizing according to the ratio between the current and new focal lengths, followed by center cropping to maintain the original image resolution.
Since we only augment focal lengths in the ascending direction as stated in~\cref{sec:aug_int}, the resized and cropped images remain within the originally observed field of view and do not introduce any unseen regions. 

\subsection{Video Pair Selection.}
During training, given a trajectory pair and its corresponding intrinsically augmented pair as in~\cref{fig:dataset_before_after} (b), 
we randomly select two trajectories among these four trajectories, using one as the source and the other as the target. 
Although we only augment focal lengths in the ascending direction, 
using augmented videos as either source or target ensures that the selected 
video pairs encompass both focal-length increase and decrease scenarios.    

\subsection{Data Statistics.}
By applying augmentations, we expand the dataset from 13{,}600 to a total of 47{,}600 videos. 
Among these, 47{,}432 videos are used for training, and 168 videos are reserved for testing, without scene overlap between the sets.

\begin{algorithm}
\small
\caption{Intrinsic Augmentation}
\begin{algorithmic}[1]
\State \textbf{Input:} Input/Output scene paths $P^{in}_{scene}, P^{out}_{scene}$, focal lengths $f_{old}, f_{new}$
\State \textbf{Output:} Augmented video dataset with updated focal length

\Procedure{ProcessVideos}{$P_{scene}^{in}$, $P_{scene}^{out}$, $f_{now}$, $f_{new}$}
    \For{$i = 1$ to $10$} \Comment{Process 10 cameras}
        \State $video \leftarrow$ Load($P_{scene}^{in}$/cam\_{$i$}.mp4)
        \State $(W_{orig}, H_{orig}) \leftarrow$ GetDimensions($video$)
        \State $(W_{new}, H_{new}) \leftarrow (\frac{f_{new}}{f_{old}} \cdot W_{orig}, \frac{f_{new}}{f_{old}} \cdot H_{orig})$
        
        \While{frame exists in $video$}
            \State $frame \leftarrow$ ReadFrame($video$)
            \State $frame \leftarrow$ Resize($frame$, $(W_{new}, H_{new})$)
            \State $frame \leftarrow$ CenterCrop($frame$, $(W_{orig}, H_{orig})$)
            \State WriteFrame($P_{scene}^{out}$/cam\_{$i$}.mp4, $frame$)
        \EndWhile
    \EndFor
\EndProcedure
\label{sup_alg:aug_focal}
\end{algorithmic}
\end{algorithm}


\section{Additional Experimental Details}
\subsection{Loss Function}
We employ the same training objective as Wan2.1~\cite{wan2025}.
Wan2.1 is an open-source Text-to-Video (T2V) diffusion model based on a transformer architecture.
During training, for a given video $\mathbf{V} \in \mathbb{R}^{B \times (1+F) \times H \times W \times 3}$, 
the Wan-VAE compresses its spatio-temporal dimensions from $(1+F, H, W)$ to $(1 + F /4, H/8, W/8)$.
Subsequent patchification further reduces the spatial resolution, yielding 
$\mathbf{z} \in \mathbb{R}^{B \times (f \times h \times w) \times d}$
, where $f = 1+F/4$, $h = H/16$, $w = W/16$.
Given a video latent $\mathbf{z}_1$, a random noise $\mathbf{z}_0 \sim \mathcal{N}(0, I)$, and a sampled timestep $t \in [0, 1]$, 
an intermediate latent $\mathbf{z}_t$ is obtained as the training input. 
Following Rectified Flows~\cite{esser2024scaling},
$\mathbf{z}_t$ is defined as a linear interpolation between $\mathbf{z}_0$ and $\mathbf{z}_1$, i.e., $\mathbf{z}_t = t\mathbf{z}_1 + (1-t)\mathbf{z}_0$.
The ground truth velocity $\mathbf{v}_t$ is $\mathbf{v}_t = \frac{d\mathbf{z}_t}{dt} = \mathbf{z}_1 - \mathbf{z}_0.$
The model is trained to predict the velocity, thus, the loss function can be formulated as the mean squared error (MSE) between the model output and $\mathbf{v}_t$,
\begin{equation}
    \mathcal{L} = \mathbb{E}_{\mathbf{z}_0, \mathbf{z}_1, c_{txt}, t} \left[ \|u(\mathbf{z}_t, c_{txt}, t; \theta) - \mathbf{v}_t\|^2 \right],
    \label{eq:loss_function}
\end{equation}
where $c_{txt}$ is the text embedding sequence, $\theta$ represents the model weights, and $u(\mathbf{z}_t, c_{txt}, t; \theta)$ denotes the output velocity predicted by the model. 

\subsection{Baseline Setup}
\subsubsection{ReCamMaster}~\cite{bai2025recammaster}. 
Following the official code and checkpoints, we generate videos at a resolution of $480 \times 832$ and a length of 81 frames. 
In addition, as described in~\cref{sec:exp_qual}, we introduce \textit{ReCamMaster w/ interp.}, which mitigates ReCamMaster’s bias to preserve the first frame of the source video by prepending eight auxiliary frames that smoothly interpolate the camera pose of the first frame toward the target pose.

\subsubsection{TrajectoryCrafter}~\cite{mark2025trajectorycrafter}. 
Following the official code and checkpoints, we generate videos at a resolution of $384 \times 672$ and a length of 49 frames.
For comparison under the 81-frame setting, we first generate a 49-frame segment, and then generate an additional 33-frame segment.
When generating the second segment, we feed the last frame of the first segment output 
as the first frame of the reprojection condition for the second segment to enforce temporal consistency at the segment boundary.
After inference, we discard the first frame of the second segment and concatenate the remaining 32 frames to the initial 49-frame segment, obtaining an 81-frame sequence with a one-frame overlap removed.
This inference-level extension alleviates the architectural limitation while preserving temporal continuity, and all other experimental settings are kept identical to those in the original paper.

\subsubsection{GEN3C}~\cite{ren2025gen3c}. 
Following the official code and checkpoints, we generate videos at a resolution of $704 \times 1280$ and a length of 121 frames. 
For the 81-frame comparison, we extend the input sequence by duplicating its final frame 40 times and appending these frames to the end of the sequence.
GEN3C is conditioned on this extended input, and the first 81 frames of the generated output are extracted and used for evaluation.

\subsection{Camera Accuracy Metric}

For RotErr (degree) and TransErr (meter) computation, ViPE~\cite{huang2025vipe} is used for camera trajectory extraction of the generated video, and the extracted trajectory is compared with the ground truth trajectory.
To ensure that camera trajectory extraction with ViPE~\cite{huang2025vipe} references the initial camera pose of the source video, 
the first frame of each source video is prepended to every generated video. 
ViPE then estimates relative poses with respect to this concatenated first frame. 
The pose estimated for this prepended frame is discarded before evaluation. 
Following~\cite{he2024cameractrl}, we compute the rotation error and translation error as
\begin{align}
RotErr &= \arccos\left( \left(\operatorname{tr}\left(\textbf{R}_{\text{pred}} \textbf{R}_{\text{gt}}^{\top}\right) - 1\right)/{2} \right), \\
TransErr &= \left\| \mathbf{t}_{\text{pred}} - \mathbf{t}_{\text{gt}} \right\|_2, 
\end{align}
where $\mathbf{R}_{\text{pred}}, \mathbf{R}_{\text{gt}} \in \mathrm{SO}(3)$ are the predicted and ground-truth rotation matrices, respectively, 
and $\mathbf{t}_{\text{pred}}, \mathbf{t}_{\text{gt}} \in \mathbb{R}^3$ are their corresponding translation vectors.

\section{Limitation Analysis}
\label{sup_sec:limit}
\begin{figure}[t]
  \centering  
  \includegraphics[width=0.9\linewidth]{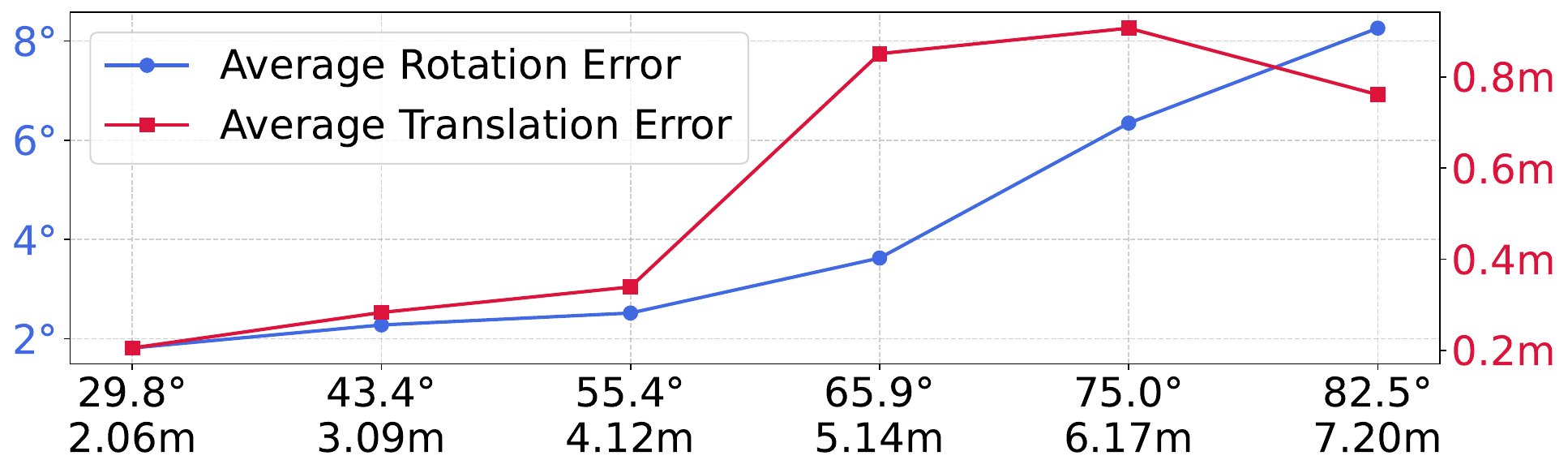}
   \caption{Camera pose accuracy under varying motion magnitudes. The X-axis represents the magnitude of rotation and translation, and the Y-axis represents the average errors.}
   \label{fig:limit}
\end{figure}
\begin{figure}[t]
    \centering
    \includegraphics[width=\linewidth]{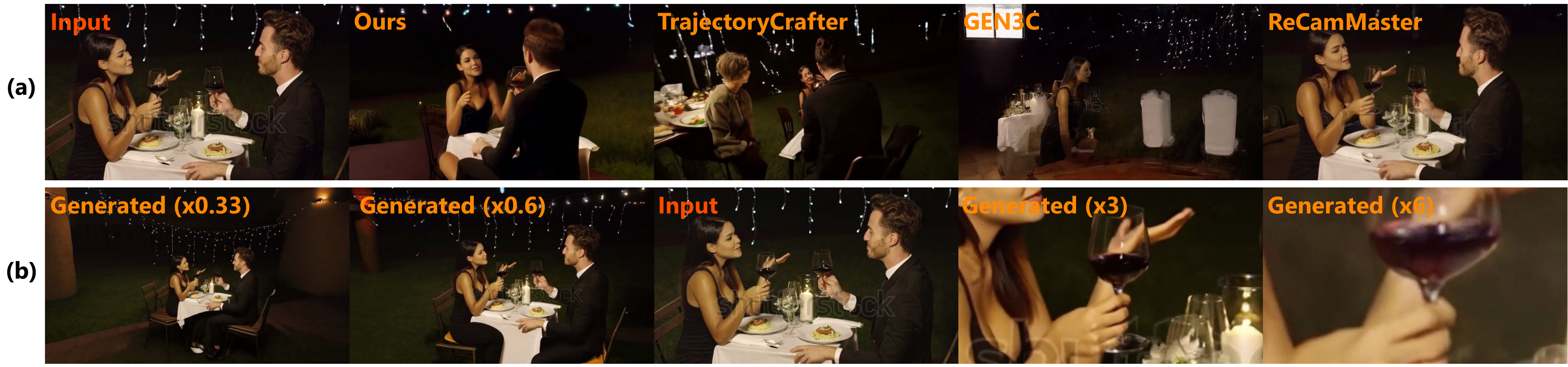}
    \caption{ Analysis of control limits. (a) Our model at extreme poses ($55.4^\circ$ rot., $4.12m$ trans.) where errors start to accumulate. At this limit, GEN3C and TrajectoryCrafter produce severe visual artifacts, and ReCamMaster fails to reflect the target motion. (b) Zoom limits: High-fidelity results are maintained up to 3$\times$ zoom, with gradual texture degradation observed at 6$\times$ magnification.}
    \label{fig:extreme}
\end{figure}
To evaluate the control limits of~\modelname{}, we conducted experiments by progressively increasing the terminal-frame rotation angle and translation magnitude for an 81-frame sequence. 

As shown in~\cref{fig:limit}, trajectory errors rise beyond $55.4^\circ$ rotation and $4.12m$ translation. 
While \modelname{} exhibits substantial robustness and extrapolation capabilities far beyond the training distribution, where our dataset's mean rotation is $10.75^\circ$ ($\sigma=11.20^\circ$) and mean translation is $0.68m$ ($\sigma=0.85m$), further refinement is required for camera poses exceeding these thresholds. 
To visualize the magnitude of camera movement represented by $55.4^\circ$ and $4.12m$, we provide qualitative comparisons of video results generated by our method and the baselines in \cref{fig:extreme}(a).
The figure displays the terminal frame of the video sequence. 
As visualized, existing state-of-the-art methods fail to produce plausible results at $55.4^\circ$ rotation and $4.12m$ translation. TrajectoryCrafter and GEN3C generate severely distorted frames, while ReCamMaster remains nearly static, failing to execute the required large-scale motion.

To evaluate the zoom capabilities, we simulated four camera categories aligned with industry-standard focal lengths: ultra-wide (0.33x), wide (0.6x), telephoto (3x), and super-telephoto (6x), defining the input focal length as the standard (1x) reference.
As visualized in~\cref{fig:extreme}(b),~\modelname{} maintains high fidelity up to 3x zoom, though 6x magnification introduces blur or certain visual artifacts.

\section{Additional Results}
\subsection{Additional Quantitative Results}
\subsubsection{Initial Frame Conditioning and Per-frame Conditioning}
\label{sup:perframe}
Our proposed method (\cref{sec:model}) primarily conditions on the warped initial-frame latent of the source video, using the target camera pose defined relative to the initial frame. 
Alternatively, one could consider a per-frame conditioning approach, where the warped latents are derived using target camera poses relative to each respective frame.
To evaluate the effectiveness of these strategies, we compare Ours-IF (Initial Frame conditioning) and Ours-PF (Per-Frame conditioning).
Evaluation is conducted on the test set of \dataname (shared intrinsic case).
\begin{table}[h]
\centering
\caption{Quantitative comparison on the ~\dataname{} test set. The best and second-best results are \textbf{bold} and \underline{underlined}, respectively.}
\label{tab:quan_indomain_pf}
\begin{tabular}{l|ccc}
\toprule
Method & PSNR$\uparrow$ & SSIM$\uparrow$ & LPIPS$\downarrow$ \\
\midrule
GEN3C~\cite{ren2025gen3c} & 16.891 & 0.479 & 0.548 \\
ReCamMaster~\cite{bai2025recammaster} & 21.130 & 0.617 & 0.420 \\
TrajectoryCrafter~\cite{mark2025trajectorycrafter} & 21.228 & 0.660 & 0.296 \\
\rowcolor{gray!10}
Ours-PF & \underline{22.231} & \underline{0.715} & \underline{0.269} \\
\rowcolor{gray!20}
Ours-IF & \textbf{22.677} & \textbf{0.718} & \textbf{0.246} \\
\bottomrule
\end{tabular}
\end{table}

As shown in \cref{tab:quan_indomain_pf}, 
both Ours-IF and Ours-PF outperform the baseline methods, demonstrating the robust architectural advantage of our Infinite-Homography conditioning regardless of the conditional strategy employed. Between the two, we further observe that Ours-IF consistently outperforms Ours-PF in terms of PSNR, SSIM, and LPIPS.

\subsubsection{Pixel-Space versus Latent-Space Warping}\label{sup_sec:warp_compare}
\begin{figure}[t]
  \centering
  \includegraphics[width=0.85\columnwidth]{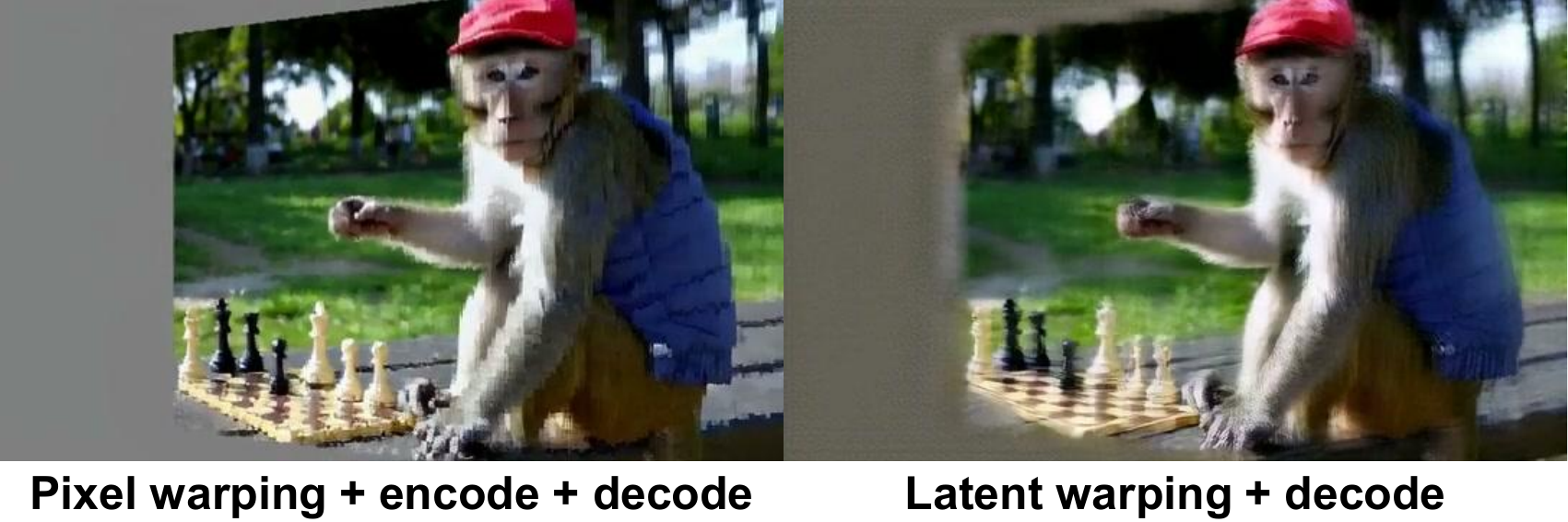}
  \caption{Left: pixel-space warp + VAE encode--decode. Right: latent-space warp + VAE decode.}
  \label{fig:latent_decode}
\end{figure}

Our final model applies infinite-homography warping directly in the latent space for efficiency, whereas warping in pixel space is a plausible alternative that may better preserve fine image details before the VAE encoding.
To examine the effect of switching to pixel-space warping, we retain our core idea of infinite-homography warping and ablate only the space in which the warping is applied.
In one variant, the input is homography-warped in pixel space, VAE-encoded, and used in place of the warped latent; in the other, the warping is applied directly in the latent space, as in our final model.
As shown in~\cref{tab:warp_space}, warping in pixel space yields slightly higher PSNR/SSIM, while warping in the latent space yields slightly better LPIPS, and the gap is marginal across all metrics.
As visualized in~\cref{fig:latent_decode}, the decoded latent-space warp is blurrier than the pixel-space warp, yet it still retains information comparable to the VAE-encoded pixel-space warp.
We presume that this comparable information content is what keeps the two variants on par.
In terms of efficiency, pixel-space warping must re-encode every warped image through the VAE encoder at each training step, whereas latent-space warping operates directly on pre-encoded latents; this makes pixel-space training about 20\% slower, so we adopt latent-space warping.

\begin{table}[t]
\centering
\scriptsize
\setlength{\tabcolsep}{3pt}
\caption{Ablation on the \dataname{} test set comparing the space in which infinite-homography warping is applied. \textit{Ours (pixel warp)} warps the input in pixel space and then VAE-encodes it, whereas \textit{Ours (latent warp)} applies the warping directly in the latent space. All models are trained for 20k steps at $320\times544$ resolution. The best and second-best results are \textbf{bold} and \underline{underlined}, respectively.}
\label{tab:warp_space}
\newcolumntype{Y}{>{\centering\arraybackslash}X}
\begin{tabularx}{\columnwidth}{l YYY YYY}
\toprule
& \multicolumn{3}{c}{Shared Intrinsic} & \multicolumn{3}{c}{Different Intrinsic} \\
\cmidrule(lr){2-4} \cmidrule(lr){5-7}
Method & PSNR$\uparrow$ & SSIM$\uparrow$ & LPIPS$\downarrow$ & PSNR$\uparrow$ & SSIM$\uparrow$ & LPIPS$\downarrow$ \\
\midrule
Ours (pixel warp)  & \textbf{24.795} & \textbf{0.738} & \underline{0.203} & \textbf{24.670} & \textbf{0.728} & \underline{0.207} \\
\addlinespace[0.5ex]
\rowcolor{gray!15}
Ours (latent warp)       & \underline{24.412} & \underline{0.733} & \textbf{0.198} & \underline{24.311} & \underline{0.720} & \textbf{0.203} \\
\midrule
ReCamMaster~\cite{bai2025recammaster}                & 20.505 & 0.589 & 0.397 & 18.377 & 0.504 & 0.524 \\
ReCamMaster (w/ \dataname{})    & 21.457 & 0.634 & 0.306 & 18.646 & 0.523 & 0.492 \\
\bottomrule
\end{tabularx}
\end{table}

\subsubsection{Random Trajectories on Prior Methods' Flagship Inputs}\label{sup_sec:depth_reproj}

\begin{figure}[!t]
  \centering
  \includegraphics[width=\columnwidth]{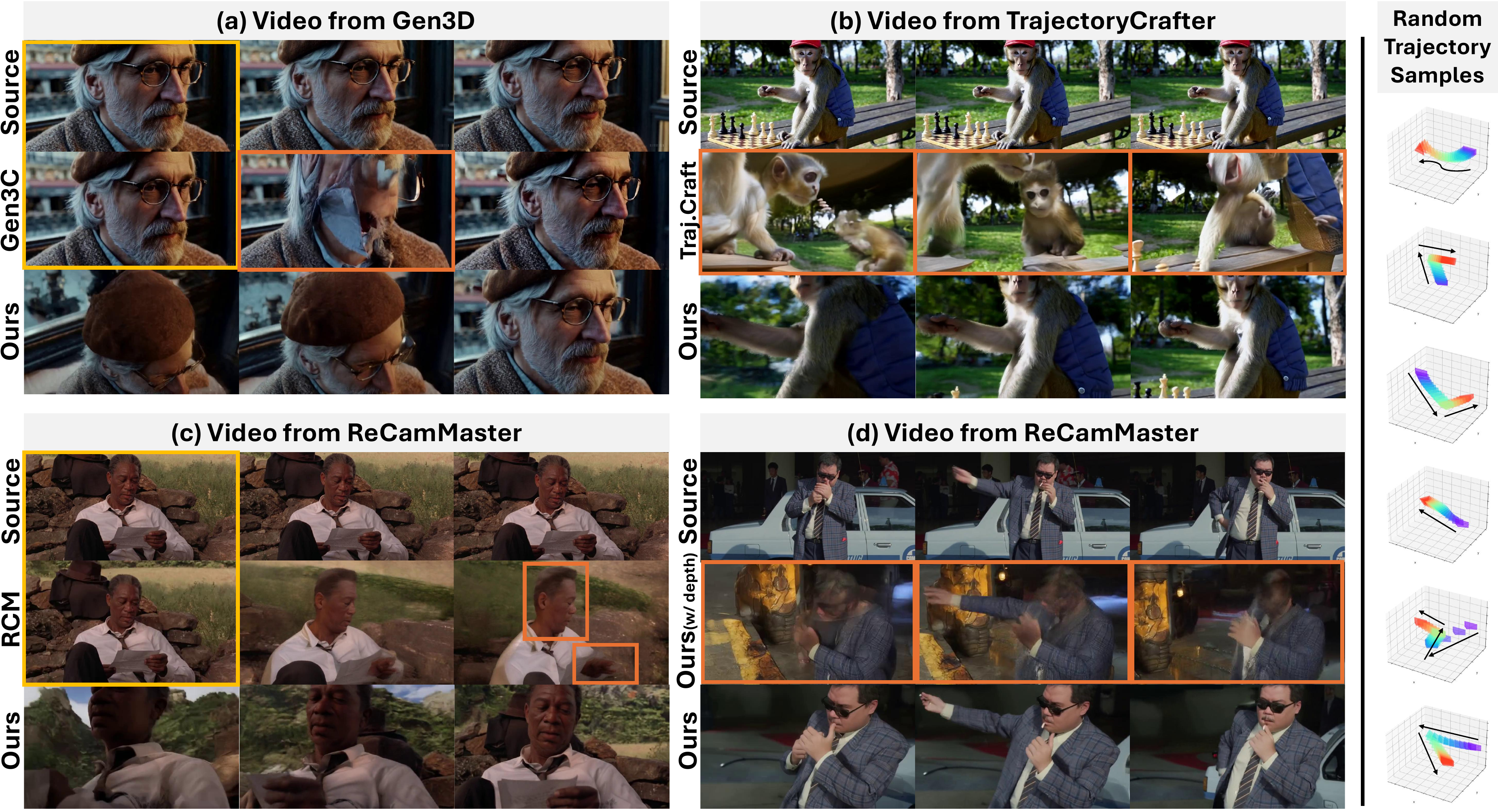}
  \captionof{figure}{The rightmost column visualizes some of the random trajectories used in evaluation; panels (b)/(c) apply trajectories of $\sim$80\textdegree{} rotation with translation. \textcolor{yellow!75!black}{Yellow} boxes mark the first frame stuck at the source pose (a typical baseline failure); \textcolor{orange}{orange} boxes mark visible artifacts.}
  \label{fig:flagship}

  \vspace{2mm}
  \captionof{table}{Quantitative comparison on demo inputs released by the baselines, with random target trajectories sampled from \dataname{}. MS denotes Motion Smoothness from VBench. The best and second-best results are \textbf{bold} and \underline{underlined}, respectively.}
  \label{tab:demo_eval}
  \scriptsize
  \setlength{\tabcolsep}{3pt}
  \newcolumntype{Y}{>{\centering\arraybackslash}X}
  \begin{tabularx}{\columnwidth}{l YYYYY}
  \toprule
  Method & Rot.Err$\downarrow$ & Trans.Err$\downarrow$ & FID$\downarrow$ & FVD$\downarrow$ & MS$\uparrow$ \\
  \midrule
  TC                  & \underline{18.818} & \underline{1.722} & \underline{97.581} & \underline{723.037} & \underline{0.985} \\
  RCM                 & 30.519 & 2.316 & 109.736 & 824.466 & \textbf{0.990} \\
  Gen3C               & 19.872 & 1.964 & 113.793 & 798.028 & 0.982 \\
  Ours (w/ depth)     & 19.572 & 1.808 & 123.519 & 806.609 & \underline{0.985} \\
  \addlinespace[0.5ex]
  \rowcolor{gray!15}
  Ours                & \textbf{14.279} & \textbf{1.540} & \textbf{91.416} & \textbf{604.478} & \textbf{0.990} \\
  \bottomrule
  \end{tabularx}
\end{figure}

We further evaluate on demo inputs released in the GitHub repositories of GEN3C~\cite{ren2025gen3c}, ReCamMaster~\cite{bai2025recammaster} (RCM), and TrajectoryCrafter~\cite{mark2025trajectorycrafter} (TC); the evaluation set comprises 25 demo inputs $\times$ 9 random trajectories sampled from \dataname{}, totaling 225 generated videos.
\modelname{} achieves the best results across all metrics in \cref{tab:demo_eval}; qualitative comparisons are shown in \cref{fig:flagship}.
We additionally evaluate a depth-based variant of our method, \textit{Ours (w/ depth)}, which makes the depth-reprojection step explicit instead of relying on our infinite-homography conditioning.
Concretely, rather than conditioning on our warped latent, we estimate per-pixel depth with UniDepth~\cite{piccinelli2024unidepth}, use it to reproject the source video to the target view in pixel space, VAE-encode the reprojected video, and use the resulting latent as the conditioning input.
This variant follows the same depth-reprojection strategy as TrajectoryCrafter (TC) and GEN3C (Gen3C).
As shown in \cref{tab:demo_eval}, \textit{Ours (w/ depth)} underperforms our final model, as do TC and Gen3C, confirming that error-prone depth reprojection is a key bottleneck of the pipeline, whereas our depth-free infinite-homography conditioning avoids it.

\subsubsection{Extended Comparison with ReCamMaster}\label{sup_sec:rcm_compare}
We note that our base model, which uses neither our \dataname{} augmentation nor infinite-homography warping and is thus trained on the original MultiCamVideo dataset, trails ReCamMaster~\cite{bai2025recammaster}, as seen by comparing the baseline configuration in \cref{tab:quan_ablation} with ReCamMaster in \cref{tab:warp_space}.
We conjecture that this gap arises because ReCamMaster conditions every output frame on \emph{all} source frames, whereas \modelname{}'s \emph{homography-guided self-attention} (\cref{fig:model}) by design conditions only on the temporally-aligned source frame, so ReCamMaster's larger conditioning context is advantageous.
However, when trained on the more complex \dataname{}, even without warping, our model (\cref{tab:quan_ablation}, Aug.Traj.+Aug.Intr.) surpasses \textit{ReCamMaster (w/ \dataname{})} in \cref{tab:warp_space}, indicating that per-frame conditioning better fits diverse trajectories.
Adding infinite-homography warping on top yields a further substantial gain in \cref{tab:quan_ablation}, producing our best results.
Thus, trajectory-rich data construction and a model design tailored to it contribute independent, complementary gains.

\begin{table}[t]
\centering
\caption{Quantitative comparison of FF-Sync case on WebVid dataset. S.C. and M.S. denote Subject Consistency and Motion Smoothness. The best results are in \textbf{bold}; the second-best are \underline{underlined}.}
\label{tab:quan_sync}
\newcolumntype{Y}{>{\centering\arraybackslash}X}

\begin{tabularx}{\columnwidth}{l|YYYYYY}
\toprule
& \multicolumn{6}{|c}{FF-Sync}\\
Method & RotErr$\downarrow$ & TransErr$\downarrow$ & FID$\downarrow$ & FVD$\downarrow$ & S.C.$\uparrow$ & M.S.$\uparrow$ \\
\midrule
GEN3C~\cite{ren2025gen3c} & 10.012 & 2.329 & 53.103 & 350.499 & 0.882 & 0.991 \\
ReCamMaster~\cite{bai2025recammaster}            & 9.673 & 1.466 & 40.612 & 308.697 & \underline{0.942} & \textbf{0.994} \\
TrajectoryCrafter~\cite{mark2025trajectorycrafter} & \underline{5.595} & \textbf{0.502} & \textbf{32.220} & \underline{287.805}  & 0.938 & 0.990 \\
\rowcolor{gray!15} 
Ours                   & \textbf{3.605} & \underline{0.510} & \underline{32.906} & \textbf{282.703} & \textbf{0.949} & \underline{0.993} \\
\bottomrule
\end{tabularx}

\vspace{1ex}

\caption{Quantitative comparison of FF-Async case on WebVid dataset. S.C. and M.S. denote Subject Consistency and Motion Smoothness.  The best results are in \textbf{bold}; the second-best are \underline{underlined}.}
\label{tab:quan_async}
\begin{tabularx}{\columnwidth}{l|YYYYYY}
\toprule
& \multicolumn{6}{|c}{FF-Async}\\
Method & RotErr$\downarrow$ & TransErr$\downarrow$ & FID$\downarrow$ & FVD$\downarrow$ & S.C.$\uparrow$ & M.S.$\uparrow$ \\
\midrule
GEN3C~\cite{ren2025gen3c} & 9.165 & 3.694 & 34.476 & 313.036 & 0.935 & 0.989 \\
ReCamMaster~\cite{bai2025recammaster}            & 7.076 & \underline{0.589} & 39.248 & \underline{295.353} & \underline{0.945} & \textbf{0.994} \\
ReCamMaster w/ interp. & 8.360 & 1.714 & \underline{30.828} & 297.328 & \textbf{0.949} & \underline{0.993} \\
TrajectoryCrafter~\cite{mark2025trajectorycrafter} & \underline{4.418} & 0.969 & 34.476 & 313.036 & 0.942 & 0.991 \\
\rowcolor{gray!15} 
Ours                   & \textbf{2.718} & \textbf{0.365} & \textbf{26.497} & \textbf{291.202} & \textbf{0.949} & \underline{0.993} \\
\bottomrule
\end{tabularx}
\end{table}

\subsubsection{FF-Sync and FF-Async Case}
\Cref{tab:quan_sync} and~\Cref{tab:quan_async} present the quantitative results for the First-Frame Synchronized (FF-Sync) and First-Frame Asynchronous (FF-Async) settings.
FF-Sync refers to the experiment setting where the source and target videos share the same initial frame, whereas FF-Async uses source and target videos with different initial frames, yielding a significantly more challenging evaluation setting.
Our method demonstrates clear superiority over competing approaches in the FF-Async setting and achieves competitive or superior performance in the FF-Sync setting.
In particular, our method attains lower rotation and translation errors in the more challenging FF-Async scenario.
These results validate that the proposed warping and augmentation strategies effectively generate videos that remain well-aligned with the target camera trajectories, regardless of the difficulty of the given trajectory configuration.
To compute the results in~\Cref{tab:quan_outdomain}, we average the scores from~\Cref{tab:quan_sync} and~\Cref{tab:quan_async}.
When performing this averaging, the value corresponding to \textit{ReCamMaster w/ interp.} in~\Cref{tab:quan_async} is paired with the value of the original ReCamMaster (without interpolation) from~\Cref{tab:quan_sync},
since interpolation is unnecessary in the synchronized case.

\subsection{Additional Qualitative Results}

\begin{figure*}[t]
    \centering
    \includegraphics[width=0.93\textwidth]{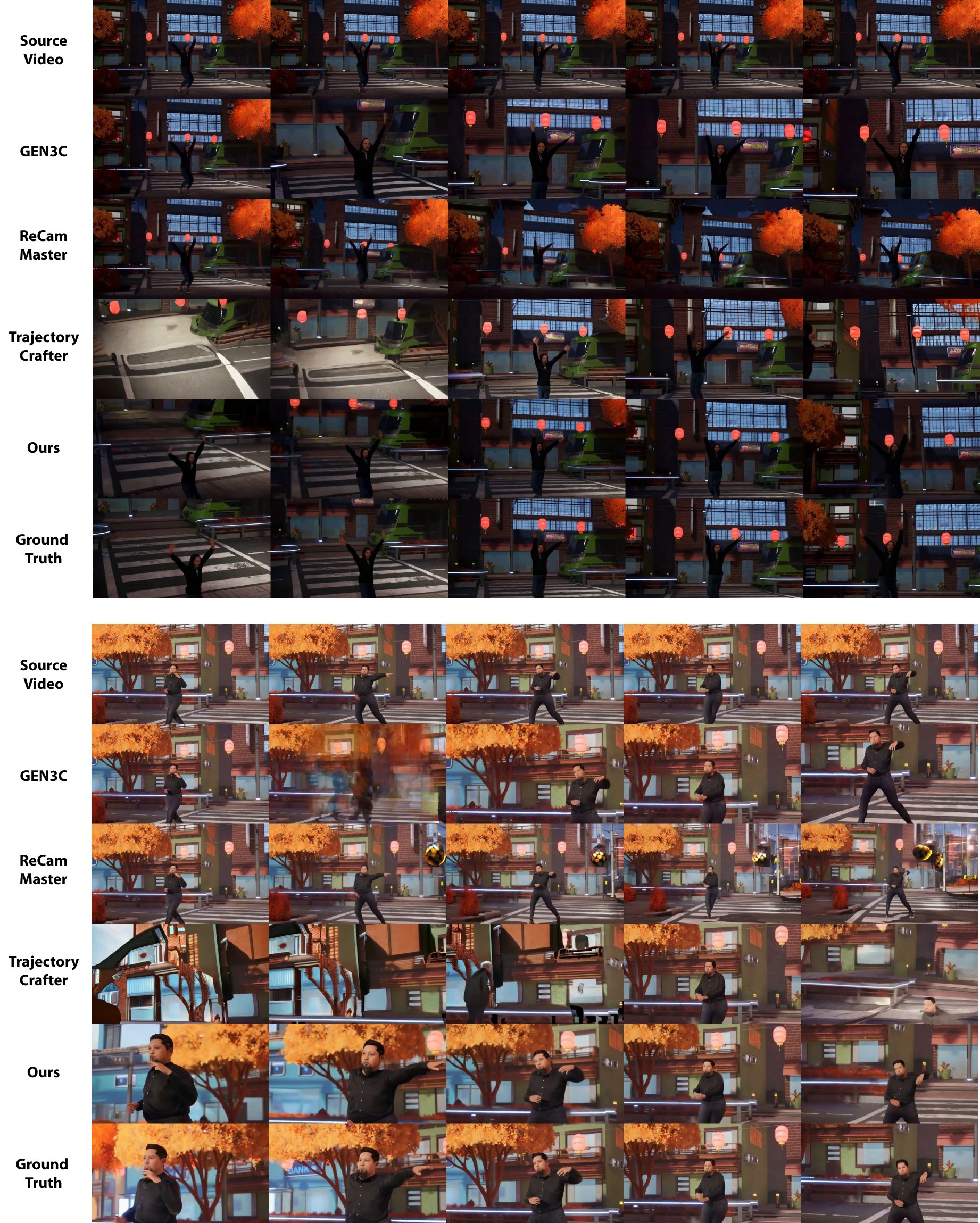}
    \caption{
    Additional qualitative results on~\dataname{} dataset. 
    Best viewed in zoom.}
    \label{fig:supple_qual_indomain}
\end{figure*}
\begin{figure*}[t]
    \centering
    \includegraphics[width=\textwidth]{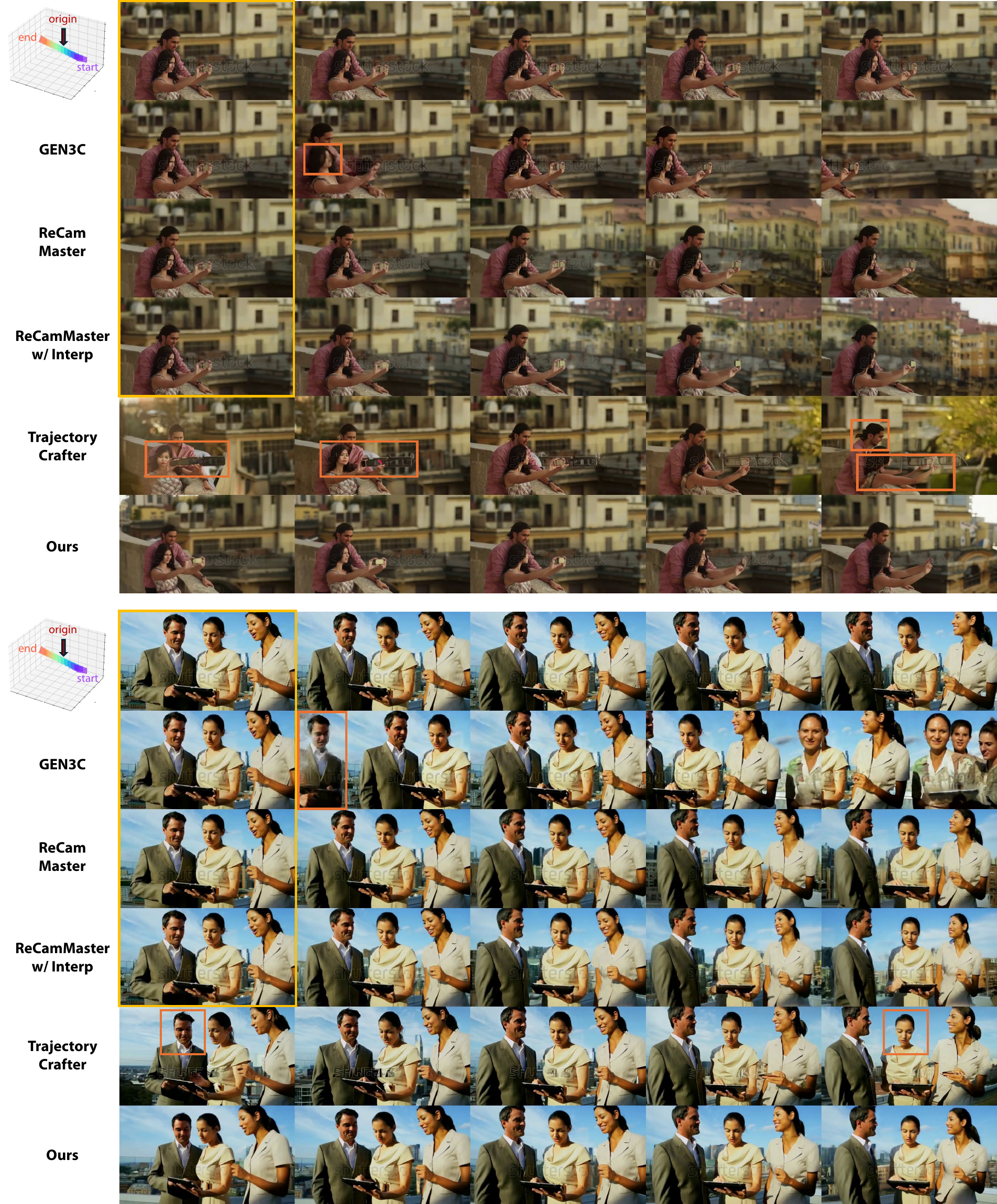}
    \caption{
    Additional qualitative results on WebVid dataset. 
    The first row illustrates the target trajectory and source video, while the remaining rows present the outputs of the baseline methods and our method. 
    Note that both GEN3C and ReCamMaster fail to modify the camera pose of the first frame (highlighted in yellow). 
    For ease of comparison, the key regions for comparison are highlighted in orange.
    Best viewed in zoom.}
    \label{fig:supple_qual_outdomain}
\end{figure*}
%
%
\begin{figure*}[t]
    \centering
    \includegraphics[width=\textwidth]{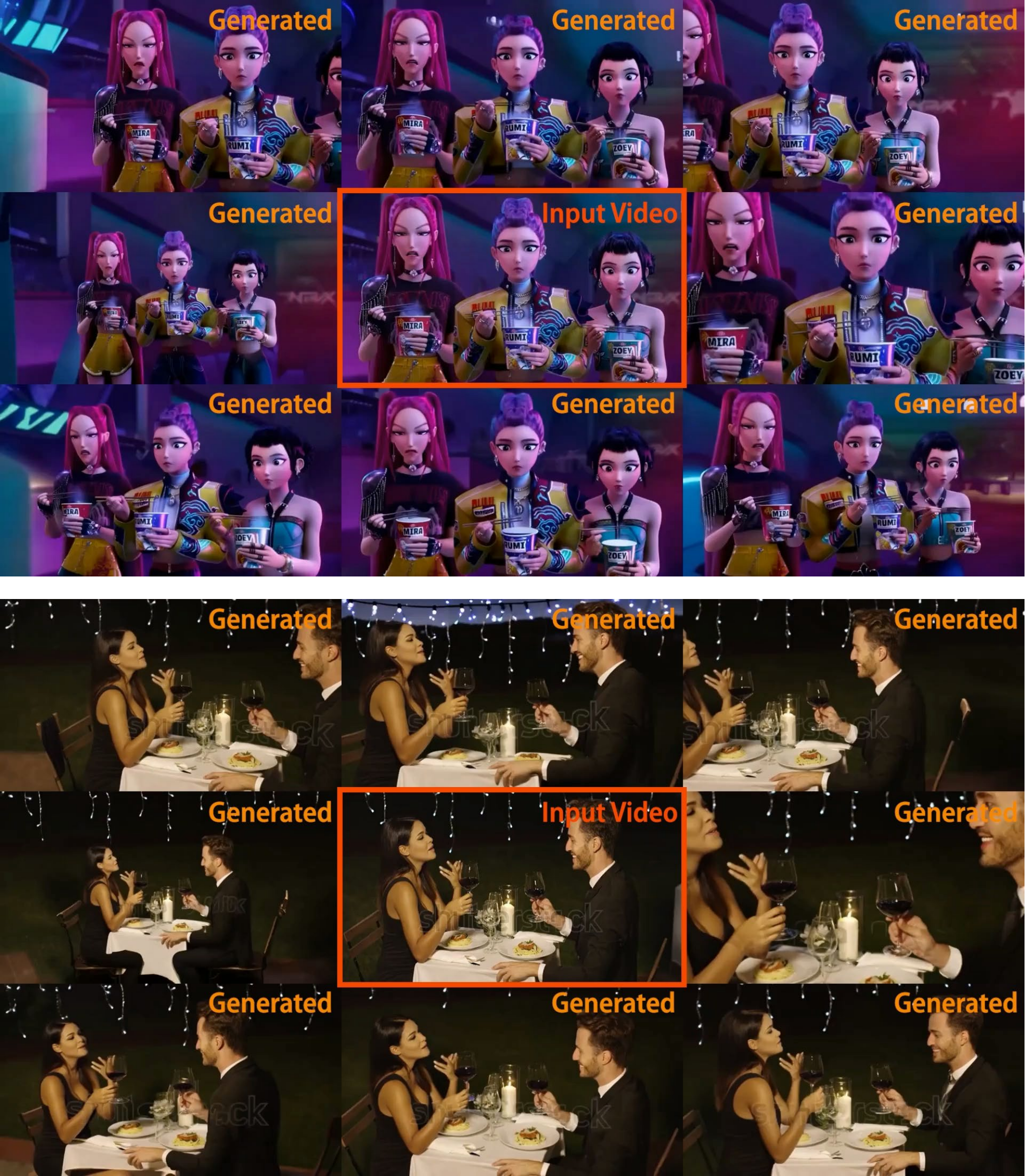}
    \caption{Additional qualitative results of our method under eight different camera trajectories. Best viewed in zoom.}
    \label{fig:supple_qual_grid}
\end{figure*}

~\cref{fig:supple_qual_indomain} and ~\cref{fig:supple_qual_outdomain} provide additional qualitative results on both the AugMCV and WebVid datasets. 
~\cref{fig:supple_qual_grid} further illustrates the generalized performance of our method across diverse camera trajectories. 
All corresponding video results are included in the supplementary material as separate video files.
